\documentclass[twoside,fleqn,hidelinks]{article}

\usepackage[l2tabu, orthodox]{nag}

\usepackage[fleqn]{amsmath}	%
\usepackage{amssymb}

\usepackage{graphicx} 		%
\usepackage{caption}
\usepackage{tabularx}
\usepackage{pbox} 			%
\usepackage{multicol, multirow}
\usepackage{tikz}
\usetikzlibrary{fit,arrows,shapes}
\usetikzlibrary{decorations.pathreplacing}

\usepackage
[
]{todonotes}
\usepackage{subcaption}
\usepackage[nolist,nohyperlinks]{acronym} %
\usepackage[utf8]{inputenc} %
\usepackage[T1]{fontenc}    %
\usepackage[english]{babel} %
\usepackage{csquotes}		%
\usepackage{hyperref}       %
\usepackage{cleveref}		%
\usepackage{url}            %
\usepackage{nicefrac}       %
\usepackage
[
activate={true,nocompatibility}, %
final,						%
tracking=true,
kerning=true,
spacing=true,
factor=1100,				%
stretch=10,					%
shrink=10					%
]
{microtype}
 \usepackage{aistats2020}

\usepackage[utf8]{inputenc} %
\usepackage[T1]{fontenc}    %
\usepackage{url}            %
\usepackage{booktabs}       %
\usepackage{array}
\usepackage{amsfonts}       %
\usepackage{nicefrac}       %
\usepackage{microtype}      %
\usepackage{tikz}

\usepackage[linesnumbered,ruled,vlined]{algorithm2e}

\usepackage{fp}
\usepackage{xifthen}

\usepackage[toc,page,header]{appendix}

\usepackage{xcolor}
\definecolor{dblue}{rgb}{0.36, 0.54, 0.66}
\usepackage{makecell}
\usepackage{mlmacros}
\usepackage[
backend=biber, %
bibencoding=utf8,%
language=auto,%
style=authoryear,%
sorting=nyt, %
maxcitenames=2,
maxbibnames=8, %
uniquelist=false,
uniquename=false,
date=year,
natbib=true %
]{biblatex}
\variables[obs,state,position,velocity,description,size,nobjects,glimpse,cnt,motion]{x,z,p,v,d,s,n,y,c,m}
\probdists{p,q}
\usepackage{twoopt}
\usepackage{scalerel}
\usepackage{moresize}
\usepackage{pgfplots}

\newcommandtwoopt{\latx}[5][\scriptsize][3.5]{{#1$#3^{#4}_{\scaleto{\textrm{#5}}{#2pt}}$}}
\newcommandtwoopt{\laty}[5][\small][4.5]{{#1$#3^{#4}_{\scaleto{\textrm{#5}}{#2pt}}$}}

\newcommand{\khead}{\bobs_{1:K}}

\newcommand{\ktail}{\bobs_{K\shortplus1:T}}

\newcommand{\mot}{MOT\xspace}
\newcommand{\find}{FIND\xspace}
\newcommand{\rect}{RECT\xspace}
\newacro{AIR}{Attend, Infer, Repeat}
\newacro{RNN}{recurrent neural network}
\newacro{CNN}{convolutional neural network}
\newacro{ours}[VTSSI]{Variational Tracking State-Space Inference}

\crefname{algocf}{algorithm}{algorithms}
\Crefname{algocf}{Algorithm}{Algorithms}
\DeclareDocumentCommand{\set}{m}{{\{#1\}}}
\usepackage{sectsty}%

\begin{document}

\twocolumn[
\title{Variational Tracking and Prediction with Generative Disentangled State-Space Models}
\aistatstitle{Variational Tracking and Prediction with Generative Disentangled State-Space Models}

\aistatsauthor{ Adnan Akhundov \And Maximilian Soelch \And Justin Bayer \And Patrick van der Smagt }
\aistatsaddress{ argmax.ai, Volkswagen Group Machine Learning Research Lab, Munich, Germany } 
]

\begin{abstract}
	We address tracking and prediction of multiple moving objects in visual data streams as inference and sampling in a disentangled latent state-space model.
	By encoding objects separately and including explicit position information in the latent state space, we perform tracking via amortized variational Bayesian inference of the respective latent positions.
	Inference is implemented in a modular neural framework tailored towards our disentangled latent space.
	Generative and inference model are jointly learned from observations only.
	Comparing to related prior work, we empirically show that our Markovian state-space assumption enables faithful and much improved long-term prediction well beyond the training horizon.
	Further, our inference model correctly decomposes frames into objects, even in the presence of occlusions.
	Tracking performance is increased significantly over prior art.
\end{abstract}
\sectionfont{\MakeUppercase}
\begin{figure*}[h!]
	\begin{subfigure}{.65\textwidth}
		\resizebox{\textwidth}{!}{
			\begin{tikzpicture}[scale=0.2]
			\path (0, -14) rectangle (59, 15);
			
			\path [draw] (4, 14) node {$\bobs$};
			\node[inner sep=0pt] at (4,9) {\includegraphics[width=8cm/5]{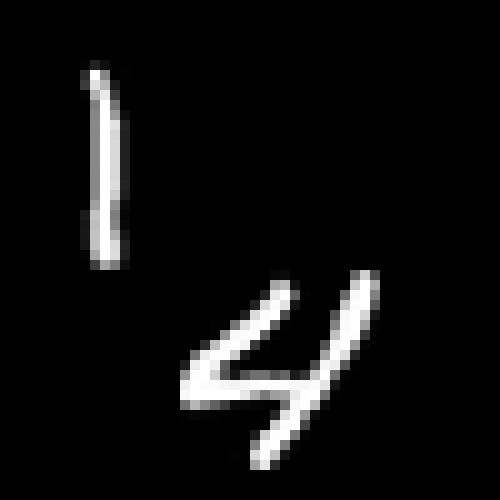}};
			
			\draw [->, >=latex, dblue, rounded corners=1] (8, 11.5) -- (14, 11.5);
			\draw [->, >=latex, dblue, rounded corners=1] (10, 11.5) -- (11, 7.5) -- (14, 7.5);
			\draw [->, >=latex, dblue, rounded corners=1] (10, 11.5) -- (11, 3.5) -- (14, 3.5);
			
			\draw [dblue, rounded corners=1] (14, 2) rectangle ++(4, 11);
			\path [draw, dblue] (16, 9) node [rotate=-90] {\Large \texttt{LSTM}};
			\draw [->, >=latex, dblue] (16, 5.5) -- (16, 3);
			
			\draw [->, >=latex, dblue] (18, 11.5) -- (22.4, 11.5);
			\draw [->, >=latex, dblue] (18, 7.5) -- (22.4, 7.5);
			\draw [->, >=latex, dblue] (18, 3.5) -- (22.4, 3.5);
			
			\path [draw] (24.5, 14) node {$\nobjects$};
			
			\path [draw] (28.5, 11.5) node {\huge [\;\;\;\;\;\;\;\;\;\;\;\;]};
			\draw [dblue, fill=green!20] (24.5, 11.5) circle [radius=1.75];
			\path (26.4, 10.2) node {\large ,};
			\draw [dblue] (28.5, 11.5) circle [radius=1.75] node [black] {$\bsize\super{1}$};
			\path (30.4, 10.2) node {\large ,};
			\draw [dblue] (32.5, 11.5) circle [radius=1.75] node [black] {$\bposition\super{1}$};
			
			\path [draw] (28.5, 7.5) node {\huge [\;\;\;\;\;\;\;\;\;\;\;\;]};
			\draw [dblue, fill=green!20] (24.5, 7.5) circle [radius=1.75];
			\path (26.4, 6.2) node {\large ,};
			\draw [dblue] (28.5, 7.5) circle [radius=1.75] node [black] {$\bsize\super{2}$};
			\path (30.4, 6.2) node {\large ,};
			\draw [dblue] (32.5, 7.5) circle [radius=1.75] node [black] 	{$\bposition\super{2}$};
			
			\path [draw] (28.5, 3.5) node {\huge [\;\;\;\;\;\;\;\;\;\;\;\;]};
			\draw [dblue, fill=red!20] (24.5, 3.5) circle [radius=1.75];
			\path (26.4, 2.2) node {\large ,};
			\draw [dblue, densely dotted] (28.5, 3.5) circle [radius=1.75] node [black] {$\bsize\super{3}$};
			\path (30.4, 2.2) node {\large ,};
			\draw [dblue, densely dotted] (32.5, 3.5) circle [radius=1.75] node [black] {$\bposition\super{3}$};			
			
			\path [draw] (42, 14) node {attended $\bobs$};
			\node[inner sep=0pt] at (42, 9) {\includegraphics[width=8cm/5]{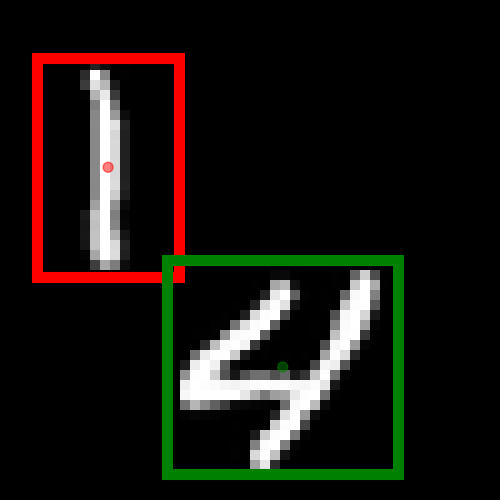}};
			
			\draw [->, >=latex, dblue] (34.6, 11.5) -- (38.55, 11.5);
			\draw [->, >=latex, dblue] (34.6, 7.5) -- (40.65, 7.5);			
			
			\draw [->, >=latex, dblue] (40.95, 10.5) -- (52, 10.5); 
			\draw [->, >=latex, dblue] (44.455, 6) -- (52, 6); 
			
			\draw [dblue, rounded corners=1] (52, 8) rectangle ++(5, 5);
			\path [draw, dblue] (54.5, 10.5) node {\Large \texttt{ST}};
			\draw [dblue, rounded corners=1] (52, 2) rectangle ++(5, 5);
			\path [draw, dblue] (54.5, 4.5) node {\Large \texttt{ST}};
			
			\draw [->, >=latex, dblue, rounded corners=1] (57, 10.5) -- (58, 10.5) -- (58, -4.5) -- (56, -4.5);
			\path [fill=white] (58, 4.5) circle [radius=0.2];
			\draw [->, >=latex, dblue, rounded corners=1] (57, 4.5) -- (59, 4.5) -- (59, -11.5) -- (56, -11.5);
			
			\path [draw] (54, -1.1) node {$\bobs\super{1}$};
			\node[inner sep=0pt] at (54, -4.5) {\includegraphics[width=4cm/5]{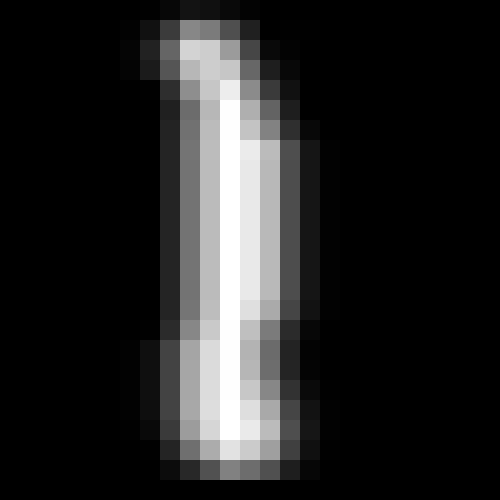}};
			\path [draw] (54, -8.1) node {$\bobs\super{2}$};
			\node[inner sep=0pt] at (54,-11.5) {\includegraphics[width=4cm/5]{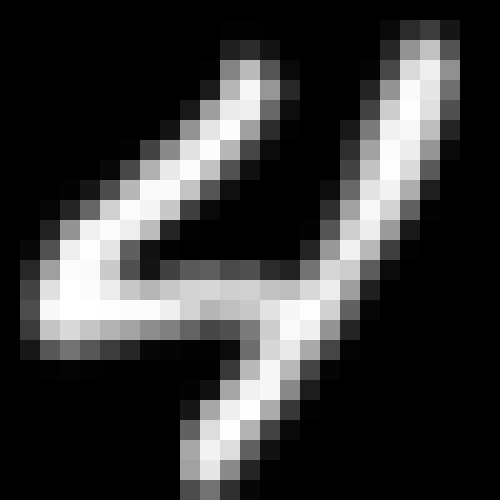}};
			
			\draw [->, >=latex, dblue] (52, -4.5) -- (48, -4.5);
			\draw [->, >=latex, dblue] (52, -11.5) -- (48, -11.5);
			
			\path [draw, dblue] (45.5, -4.5) node {\Large \texttt{VAE}};
			\draw [dblue, rounded corners=1] (48, -2) -- (48, -7) -- (43, -6) -- (43, -3) -- cycle;
			\path [draw, dblue] (45.5, -11.5) node {\Large \texttt{VAE}};
			\draw [dblue, rounded corners=1] (48, -9) -- (48, -14) -- (43, -13) -- (43, -10) -- cycle;
			
			\draw [->, >=latex, dblue] (43, -4.5) -- (41.75, -4.5);
			\draw [->, >=latex, dblue] (43, -11.5) -- (41.75, -11.5);
			
			\draw [densely dotted] (40, 1) -- (40, -14);
			\path [draw] (39.7, 0) node {\scriptsize generative \,\,\,\,\,\, inference};
			\draw [->, >=latex] (33.7, -1) -- (31.7, -1);
			\path [draw] (39.9, -1) node {\scriptsize model \;\;\;\;\;\;\;\; model};
			\draw [->, >=latex] (46, -1) -- (48, 0-1);
			
			\draw [dblue, fill=white] (40, -4.5) circle [radius=1.75] node [black] {$\bdescription\super{1}$};
			\draw [dblue, fill=white] (40, -11.5) circle [radius=1.75] node [black] {$\bdescription\super{2}$};
			
			\draw [->, >=latex, dblue] (38.25, -4.5) -- (37, -4.5);
			\draw [->, >=latex, dblue] (38.25, -11.5) -- (37, -11.5);
			
			\path [draw, dblue] (34.5, -4.5) node {\Large \texttt{VAE}};
			\draw [dblue, rounded corners=1] (32, -2) -- (32, -7) -- (37, -6) -- (37, -3) -- cycle;
			\path [draw, dblue] (34.5, -11.5) node {\Large \texttt{VAE}};
			\draw [dblue, rounded corners=1] (32, -9) -- (32, -14) -- (37, -13) -- (37, -10) -- cycle;
			
			\draw [->, >=latex, dblue] (32, -4.5) -- (28, -4.5);
			\draw [->, >=latex, dblue] (32, -11.5) -- (28, -11.5);
			
			\path [draw] (26, -1.1) node {$\bglimpse\super{1}$};
			\node[inner sep=0pt] at (26,-4.5) {\includegraphics[width=4cm/5]{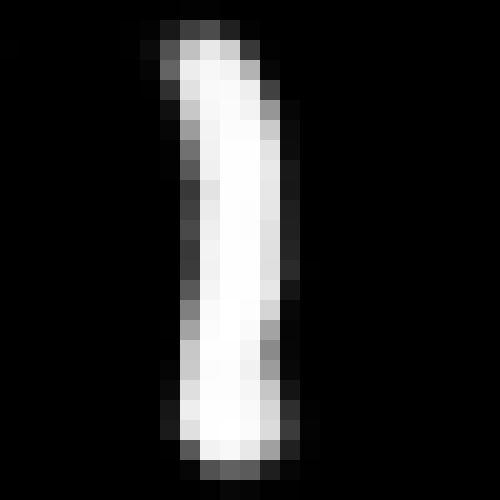}};
			\path [draw] (26, -8.1) node {$\bglimpse\super{2}$};
			\node[inner sep=0pt] at (26,-11.5) {\includegraphics[width=4cm/5]{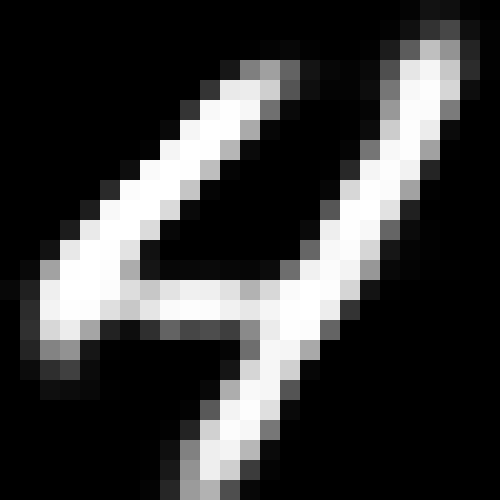}};
			
			\draw [->, >=latex, dblue] (24, -4.5) -- (19, -4.5);
			\draw [->, >=latex, dblue] (24, -11.5) -- (19, -11.5);
			
			\draw [dblue, rounded corners=1] (14, -7.5) rectangle ++(5, 5);
			\path [draw, dblue] (16.5, -4.9) node {\Large \texttt{ST$^{\texttt{-1}}$}};
			\draw [dblue, rounded corners=1] (14, -13.5) rectangle ++(5, 5);
			\path [draw, dblue] (16.5, -10.9) node {\Large \texttt{ST$^{\texttt{-1}}$}};
			
			\draw [dblue] (14, -5) -- (12, -5);
			\draw [dblue] (14, -9.5) -- (12, -9.5);
			
			\draw [dblue] (11, -5) circle [radius=1];
			\draw [dblue, thick] (10.5, -5) -- (11.5, -5);
			\draw [dblue, thick] (11, -5.5) -- (11, -4.5);
			\draw [dblue] (11, -9.5) circle [radius=1];
			\draw [dblue, thick] (10.5, -9.5) -- (11.5, -9.5);
			\draw [dblue, thick] (11, -9) -- (11, -10);
			
			\path [draw] (4, -1.5) node {reconst. $\bobs$};
			\node[inner sep=0pt] at (4,-6.5) {\includegraphics[width=8cm/5]{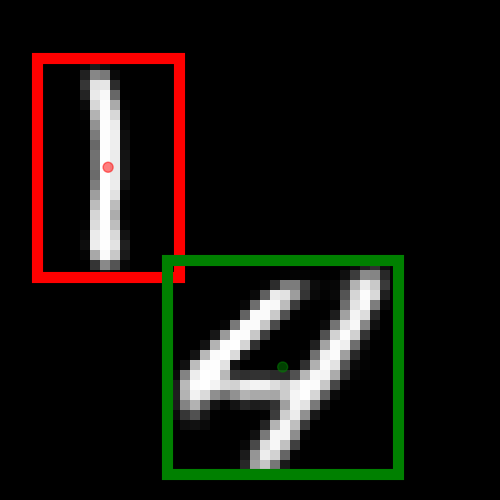}};
			
			\draw [->, >=latex, dblue] (10, -5) -- (2.9, -5); 
			\draw [->, >=latex, dblue] (10, -9.5) -- (6.4, -9.5); 				
			\end{tikzpicture}}
		
		\caption[The architecture of AIR]{The computational architecture of \ac{AIR}. \texttt{ST} and \texttt{ST}$^{\texttt{-1}}$ denote a Spatial Transformer and its inverse, \texttt{VAE} refers to a variational autoencoder.}
		\label{fig:air-architecture}
	\end{subfigure}
	\quad
	\begin{subfigure}{0.3\textwidth}
		\centering
		\resizebox{.8\textwidth}{!}{
			\begin{tikzpicture}[scale=0.2]
			\path (27, 1) rectangle (56, 20);
			
			\node[inner sep=0pt] at (30.5, 13.5) {\includegraphics[width=7cm/5]{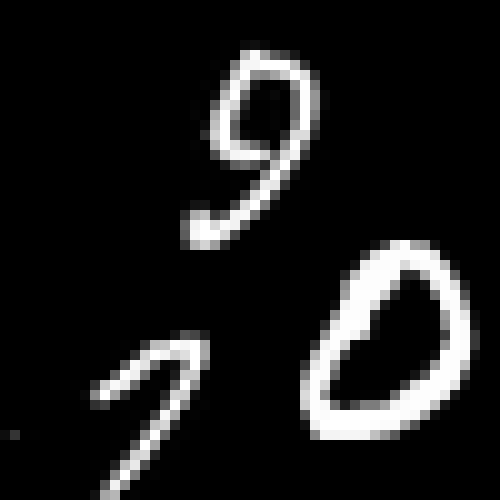}};
			
			\draw [->, >=latex, dblue] (30.5, 10) -- (30.5, 8.5);
			
			\path [draw, dblue] (30.5, 7) node {\Large \texttt{CNN}};
			\draw [dblue, rounded corners=1] (28, 5.5) rectangle ++(5, 3);
			
			\draw [->, >=latex, dblue] (30.5, 5.5) -- (30.5, 4);
			
			\draw [dblue] (30.5, 2.5) circle [radius=1.5] node [black] {\Large $\cnt$};

			\draw [dblue] (32, 2.5) -- (34, 2.5);
			\draw [->, >=latex, dblue] (34, 2.5) -- (38, 15);
			\draw [->, >=latex, dblue] (34, 2.5) -- (38, 9);
			\draw [->, >=latex, dblue] (34, 2.5) -- (38, 2.5);
			
			\draw [dblue] (38, 13) rectangle ++(4, 4) node [black, pos=0.5] {\Large $1.0$};
			\draw [dblue] (38, 7) rectangle ++(4, 4) node [black, pos=0.5] {\Large $1.0$};
			\draw [dblue] (38, 1) rectangle ++(4, 4) node [black, pos=0.5] {\Large $0.4$};
			
			\path [draw] (43.5, 15) node {\Large $\times$};
			\path [draw] (43.5, 9) node {\Large $\times$};
			\path [draw] (43.5, 3) node {\Large $\times$};
			
			\node[inner sep=0pt] at (47, 15) {\includegraphics[width=4cm/5]{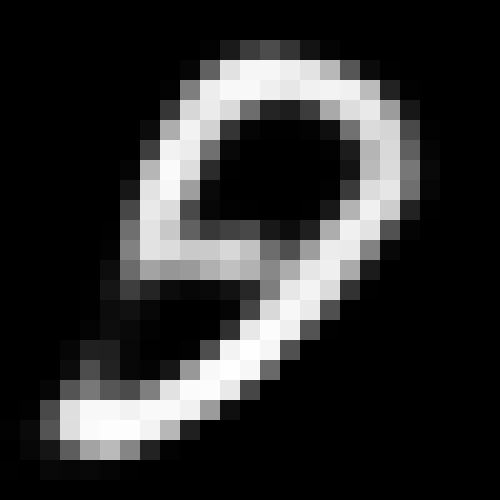}};
			\node[inner sep=0pt] at (47, 9) {\includegraphics[width=4cm/5]{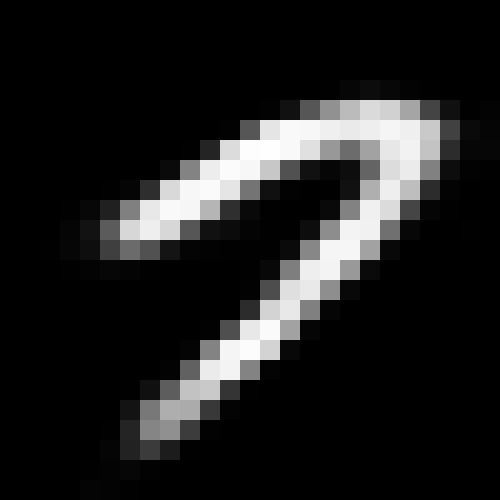}};
			\node[inner sep=0pt] at (47, 3) {\includegraphics[width=4cm/5]{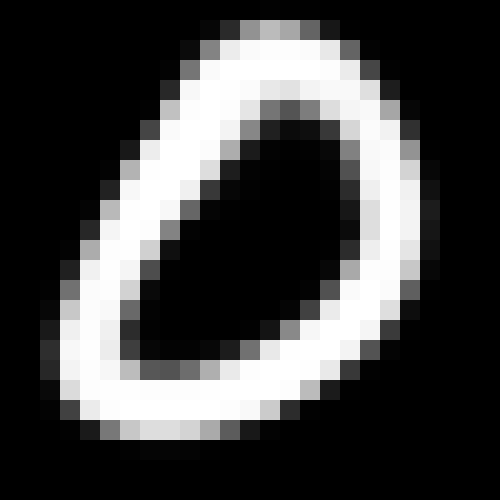}};
			
			\path [draw] (50.5, 14.8) node {\Large $=$};
			\path [draw] (50.5, 8.8) node {\Large $=$};
			\path [draw] (50.5, 2.8) node {\Large $=$};
			
			\node[inner sep=0pt] at (54, 15) {\includegraphics[width=4cm/5]{gfx/air/mnist-count-rg-1}};
			\node[inner sep=0pt] at (54, 9) {\includegraphics[width=4cm/5]{gfx/air/mnist-count-rg-2}};
			\node[inner sep=0pt] at (54, 3) {\includegraphics[width=4cm/5]{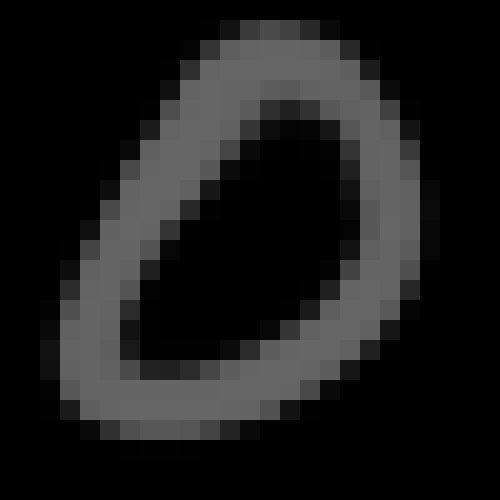}};
			
			\end{tikzpicture}
		}	\caption{Continuous counting.}
		\label{subfig:continuous-counting}
		
		\resizebox{.8\textwidth}{!}{
			\begin{tikzpicture}[scale=0.1]
			\path (1, -12) rectangle (32, 8);

			\node[inner sep=0pt] at (4, 3) {\includegraphics[width=6cm/10]{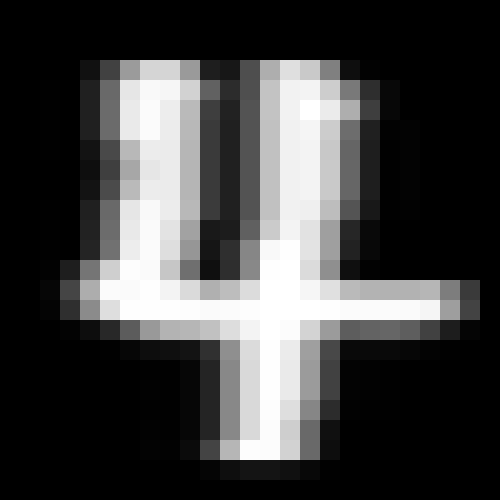}};
			
			\path [draw] (8.5, 3) node {$\times$};

			\node[inner sep=0pt] at (0.85, 0) {
				\begin{axis}[height=2.18cm, width=2.18cm, colormap/jet, view={0}{90},
				hide axis,  inner sep=0pt, ticks=none, xticklabels={,,}, yticklabels={,,}]
				\addplot3[surf,shader=flat, domain=1:25, samples=25] file {gfx/air/kernel.dat};
				\end{axis}};
			
			\path [draw] (17.5, 2.8) node {$=$};

			\node[inner sep=0pt] at (22, 3) {\includegraphics[width=6cm/10]{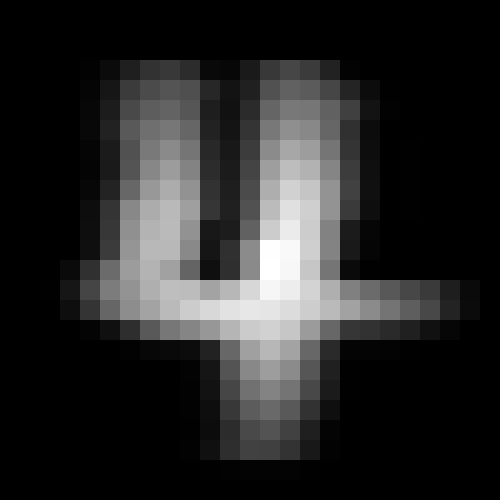}};
			
			\node[inner sep=0pt] at (6, -7) {\includegraphics[width=6cm/6]{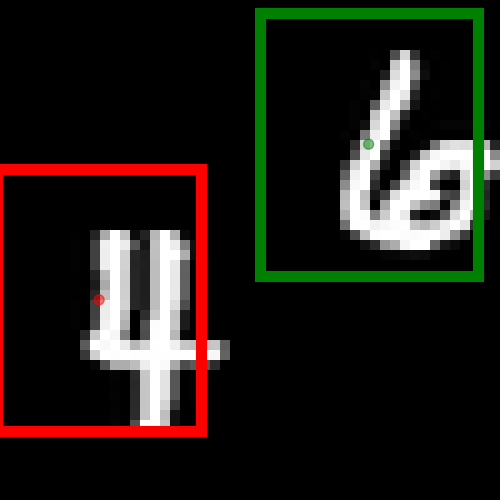}};
			\node[inner sep=0pt] at (20, -7) {\includegraphics[width=6cm/6]{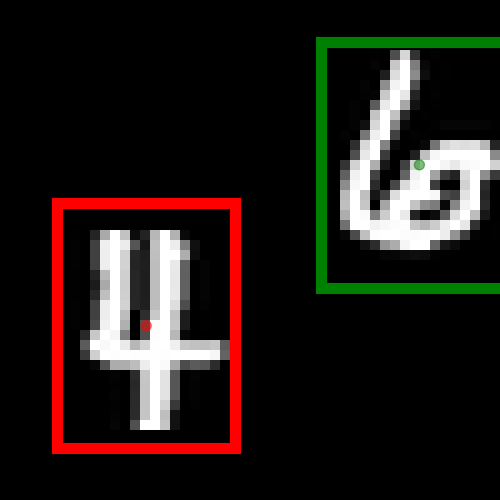}};
			\end{tikzpicture}}
		\caption{Position regularization.}			
		\label{subfig:centering-bounding-boxes}
	\end{subfigure}
	\caption{\acf{AIR}. Vanilla architecture (left) and modifications in this work (right).}
	\label{fig:air-modifications}
\end{figure*}

\section{Introduction}\label{sec:introduction}
Perception of the present and prediction of the future are key requirements for the deployment of autonomous systems in the physical world.
Many relevant and concrete perception tasks can be solved given sufficient engineering efforts \parencite{pulford_taxonomy_2005,slamsurvey}.
Adaptation of conceptually simple frameworks to specific scenarios requires the exploitation of constraints to achieve satisfying performance.
In tracking, \eg, different target representations (point, bounding box), observations (depth, color), and partial models (appearance, motion) need to be incorporated. 

In recent years, learning methods and in particular deep neural networks have enhanced or even replaced hand-crafted perception pipelines, promising competitive performance in the presence of rich data sets.
These approaches can loosely be put into three categories. 
First, components of existing pipelines are replaced by neural components, leaving major parts untouched \parencite{DBLP:conf/cvpr/SchulterVCC17,DBLP:conf/iccv/DosovitskiyFIHH15,DBLP:conf/eccv/YangWSC18}.
Second, complete pipelines are replaced with learnable counterparts, often inspired by the previously dominant solutions \parencite{DBLP:conf/nips/KrizhevskySH12,DBLP:conf/nips/KosiorekBP17,DBLP:conf/cvpr/ParisottoCZS18,DBLP:journals/ral/GordonFF18,DBLP:conf/cvpr/KahouMMPV17}.
Third, the data generating process is formulated as a latent variable model and the task of interest expressed as Bayesian inference.

The benefit of the latter is the principled quantification of uncertainty, inclusion of domain knowledge and the applicability of unsupervised and semi-supervised learning algorithms \parencite{eslami_attend_2016,DBLP:journals/corr/abs-1805-07206}.
Our work places itself in this category:
we tackle multiple-object tracking as approximate Bayesian inference in variational state-space models \parencite{DBLP:journals/corr/KrishnanSS15,archer2015black,DBLP:conf/nips/FraccaroSPW16,DBLP:conf/iclr/KarlSBS17},
a class of models that provides efficient latent representations of sequences of observations.

We adopt \acf{AIR}, a model for scene decomposition into disentangled objects. 
Our contributions are the following:
\begin{enumerate}
\item 
We modify and stabilize \ac{AIR} and extend it to sequences by adding state-space dynamics.
\item We derive an inference algorithm, \ac{ours}, to reflect the extended generative model.
\ac{ours} explicitly and efficiently exploits temporal consistency.
\item We verify that our model significantly improves  tracking and prediction performance compared to original \ac{AIR}, as well as two related baselines.
 Our model is able to decompose objects even in challenging scenarios where objects overlap.
\end{enumerate}
Overall, \ac{ours} provides a flexible, more interpretable framework for multi-object tracking and prediction.
The proposed models converge much faster with significant performance gains over state-of-the-art baselines.

\section{Attend, Infer, Repeat}\label{sub:air}

\Textcite{eslami_attend_2016} introduced \acf{AIR}, a structured variational autoencoder (VAE) for scene understanding.
In contrast to the original VAE \parencite{DBLP:journals/corr/KingmaW13,DBLP:conf/icml/RezendeMW14} it imposes structure on the generative latent-variable model:
it assumes scenes of $\nobjects\in\NN_0$ conceptually similar objects defined by a set of properties $\bstate\super{i} = \{\bposition\super{i},\bsize\super{i},\bdescription\super{i}\}$, comprised of the position $\bposition\in\RR^2$, size of the object $\bsize\in\RR^2$, and a content description vector $\bdescription\in\RR^d$.

\Cref{fig:air-architecture} illustrates inference and generation in \ac{AIR}.
During inference, an LSTM \parencite{hochreiter_long_1997} determines the amount of objects $\nobjects$ (implemented as a sequence of binary decisions), positions $\bposition\super{i}$, and extents $\bsize\super{i}$ of objects in a canvas $\bobs\in\RR^{x\times x}$.
Objects are cropped and resized to a glimpse $\bobs\super{i}\in\RR^{y\times y}$ of fixed extent via a spatial transformer \parencite{jaderberg_spatial_2015-1}.
The resulting glimpses are fed into a VAE-style encoder to obtain a fixed-size description vector $\bdescription\super{i}$.
During generation, the $\bdescription\super{i}$ are decoded into fixed-sized glimpses $\bglimpse\super{i}\in\RR^{y\times y}$.
An inverse spatial transformer  conditioned on size~$\bsize\super{i}$ and position~$\bposition\super{i}$ pastes the glimpse back to an empty scene.
By summing over all sets $\bstate\super{i}$, we obtain the full scene.
The model is trained by stochastic gradient descent on the evidence lower bound (ELBO; \cite{DBLP:journals/ml/JordanGJS99}).

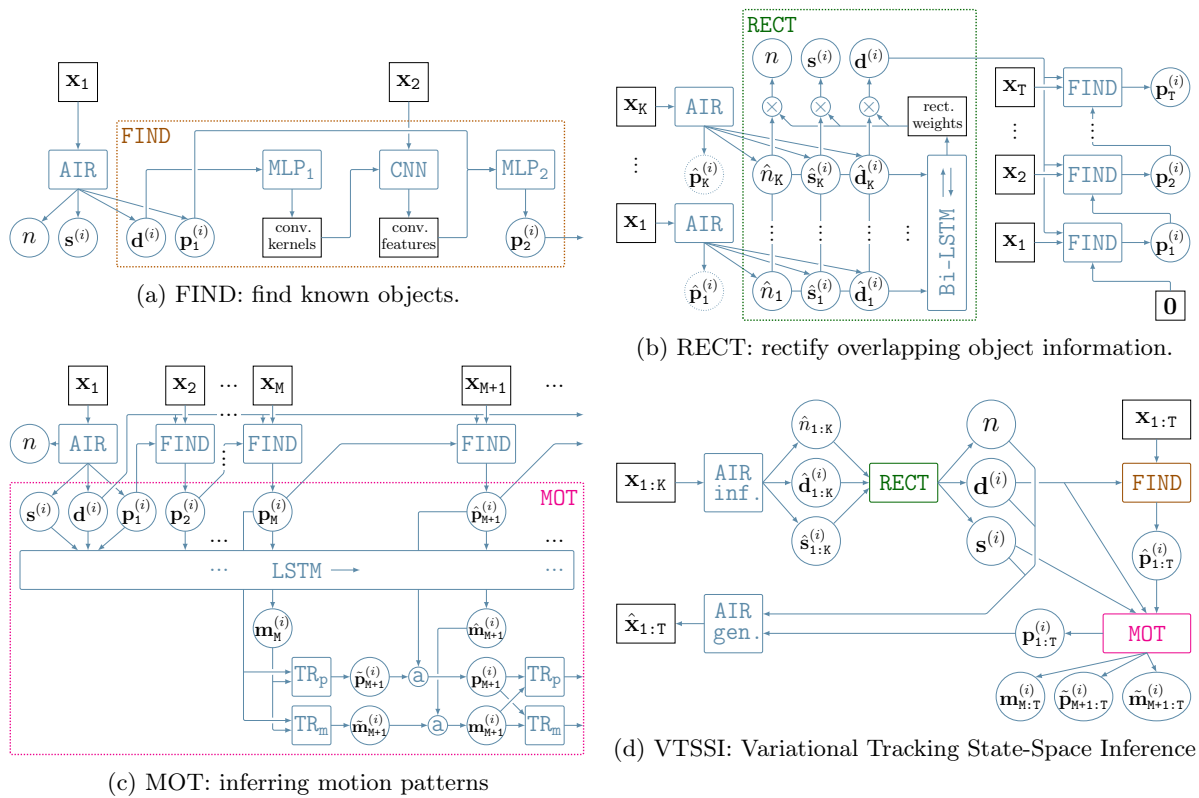
\begin{figure*}[h!]
	\begin{subfigure}{0.45\textwidth}
		\centering
		\resizebox{\textwidth}{!}{%
			\begin{tikzpicture}[scale=0.2]
				\path (-1, -10) grid (58, 11);
				
				\draw (4, 7) rectangle ++(4, 4) node [pos=0.5] {\Large $\bobs_{\texttt{1}}$};
				\draw [->, >=latex, dblue] (6, 7) -- (6, 2);
				
				\draw [dblue, rounded corners=1] (3, -2) rectangle ++(6, 4);
				\path [draw, dblue] (6, 0) node {\Large \texttt{AIR}};
				
				\draw [->, >=latex, dblue] (6, -2) -- (2.2, -5.4);
				\draw [->, >=latex, dblue] (6, -2) -- (6, -5);
				\draw [->, >=latex, dblue] (6, -2) -- (12, -5.3);
				\draw [->, >=latex, dblue] (6, -2) -- (17.35, -5.1);
				
				\draw [dblue] (1, -7) circle [radius=2] node [black] {\Large $\nobjects$};
				\draw [dblue] (6, -7) circle [radius=2] node [black] {\large $\bsize\super{i}$};
				\draw [dblue] (13, -7) circle [radius=2] node [black] {\large $\bdescription\super{i}$};
				\draw [dblue] (18, -7) circle [radius=2] node [black] {\large $\bposition\super{i}_{\texttt{1}}$};
				
				\path [fill=white] (13, -3.9) circle [radius=0.3];
				\draw [->, >=latex, dblue, rounded corners=1] (13, -5) -- (13, 0) -- (25, 0);
				
				\draw (38, 7) rectangle ++(4, 4) node [pos=0.5] {\Large $\bobs_{\texttt{2}}$};
				\draw [->, >=latex, dblue] (40, 7) -- (40, 2);
				
				\path [fill=white] (18, 0) circle [radius=0.3];
				\path [fill=white] (40, 4) circle [radius=0.3];
				\draw [->, >=latex, dblue, rounded corners=1] (18, -5) -- (18, 4) -- (46, 4) -- (46, 0) -- (49, 0);
				
				\draw [dblue, rounded corners=1] (25, -2) rectangle ++(6, 4);
				\path [draw, dblue] (28, 0) node {\Large $\texttt{MLP}_{\texttt{1}}$};
				\draw [dblue, rounded corners=1] (37, -2) rectangle ++(6, 4);
				\path [draw, dblue] (40, 0) node {\Large \texttt{CNN}};
				\draw [dblue, rounded corners=1] (49, -2) rectangle ++(6, 4);
				\path [draw, dblue] (52, 0) node {\Large $\texttt{MLP}_{\texttt{2}}$};
				
				\draw [->, >=latex, dblue] (28, -2) -- (28, -5);
				\draw (25, -9) rectangle ++(6, 4) node [text width=1cm, align=center, pos=0.5, font=\small\linespread{1.0}\selectfont] {conv. kernels};
				\draw[->, >=latex, dblue, rounded corners=1] (31, -7) -- (34, -7) -- (34, 0) -- (37, 0);
				
				\draw [->, >=latex, dblue] (40, -2) -- (40, -5);
				\draw (37, -9) rectangle ++(6, 4) node [text width=1.2cm, align=center, pos=0.5, font=\small\linespread{1.0}\selectfont] {conv. features};
				\draw[->, >=latex, dblue, rounded corners=1] (43, -7) -- (46, -7) -- (46, 0) -- (49, 0);
				
				\draw [->, >=latex, dblue] (52, -2) -- (52, -5);
				\draw [dblue] (52, -7) circle [radius=2] node [black] {\large $\bposition\super{i}_{\texttt{2}}$};
				
				\draw [->, >=latex, dblue] (54, -7) -- (58, -7);
				
				\draw [thick, densely dotted, black!30!orange, rounded corners=1] (10, -10) rectangle (56, 5);
				\path [draw, black!40!orange] (13, 3.5) node {\Large \texttt{FIND}};
			\end{tikzpicture} %
}
		\caption{\find: find known objects.}
		\label{fig:arch-AIR-FIND}
	\end{subfigure}\quad
	\begin{subfigure}{0.45\textwidth}
		\centering
		\resizebox{\textwidth}{!}{%
\begin{tikzpicture}[scale=0.2]
				\path (-3, -10) grid (56, 22);
				
				\draw (-3, -2) rectangle ++(4, 4) node [pos=0.5] {\Large $\bobs_{\texttt{1}}$};
				\draw [->, >=latex, dblue] (1, 0) -- (3, 0);
				\path [draw] (-1, 6) node [rotate=90] {\Large ...};				
				\draw (-3, 10) rectangle ++(4, 4) node [pos=0.5] {\Large $\bobs_{\texttt{K}}$};
				\draw [->, >=latex, dblue] (1, 12) -- (3, 12);
				
				\draw [dblue, rounded corners=1] (3, -2) rectangle ++(6, 4);
				\path [draw, dblue] (6, 0) node {\Large \texttt{AIR}};
				\draw [->, >=latex, dblue] (6, -2) -- (6, -5);
				\draw [->, >=latex, dblue] (6, -2) -- (11.5, -5.7);
				\draw [->, >=latex, dblue] (6, -2) -- (17, -5.3);
				\draw [->, >=latex, dblue] (6, -2) -- (22.35, -5.1);
				
				\draw [dblue, densely dotted] (6, -7) circle [radius=2] node [black] {\large $\hat{\bposition}\super{i}_{\texttt{1}}$};
				\draw [dblue] (13, -7) circle [radius=2] node [black] {\Large $\hat{\nobjects}_{\texttt{1}}$};
				\draw [dblue] (18, -7) circle [radius=2] node [black] {\large $\hat{\bsize}\super{i}_{\texttt{1}}$};
				\draw [dblue] (23, -7) circle [radius=2] node [black] {\large $\hat{\bdescription}\super{i}_{\texttt{1}}$};
				
				\draw [dblue, rounded corners=1] (3, 10) rectangle ++(6, 4);
				\path [draw, dblue] (6, 12) node {\Large \texttt{AIR}};		
				\draw [->, >=latex, dblue] (6, 10) -- (6, 7);
				\draw [->, >=latex, dblue] (6, 10) -- (11.5, 6.3);
				\draw [->, >=latex, dblue] (6, 10) -- (17, 6.7);
				\draw [->, >=latex, dblue] (6, 10) -- (22.35, 6.9);
				
				\draw [dblue, densely dotted] (6, 5) circle [radius=2] node [black] {\large $\hat{\bposition}\super{i}_{\texttt{K}}$};
				\draw [dblue] (13, 5) circle [radius=2] node [black] {\Large $\hat{\nobjects}_{\texttt{K}}$};
				\draw [dblue] (18, 5) circle [radius=2] node [black] {\large $\hat{\bsize}\super{i}_{\texttt{K}}$};
				\draw [dblue] (23, 5) circle [radius=2] node [black] {\large $\hat{\bdescription}\super{i}_{\texttt{K}}$};
				
				\draw [dblue] (15, 5) -- (16, 5);
				\draw [dblue] (20, 5) -- (21, 5);
				\draw [->, >=latex, dblue] (25, 5) -- (29, 5);
				\path [draw] (27, -1) node [rotate=90] {\Large ...};
				\draw [dblue] (15, -7) -- (16, -7);
				\draw [dblue] (20, -7) -- (21, -7);
				\draw [->, >=latex, dblue] (25, -7) -- (29, -7);
				
				\draw [dblue, rounded corners=1] (29, -9) rectangle ++(4, 16);
				\path [draw, dblue] (31, -3) node [rotate=90] {\Large \texttt{Bi-LSTM}};
				\draw [->, >=latex, dblue] (30.6, 2.7) -- (30.6, 5.7);
				\draw [<-, >=latex, dblue] (31.4, 2.7) -- (31.4, 5.7);
				
				\path [fill=white] (13, -4.2) circle [radius=0.3];
				\path [fill=white] (13, -3.4) circle [radius=0.3];
				\draw [dblue]  (13, -5) -- (13, 3) node [black, fill=white, pos=0.5, inner sep=1.0, rotate=90] {\Large ...};
				\path [fill=white] (18, -4.4) circle [radius=0.3];
				\draw [dblue] (18, -5) -- (18, 3) node [black, fill=white, pos=0.5, inner sep=1.0, rotate=90] {\Large ...};
				\draw [dblue] (23, -5) -- (23, 3) node [black, fill=white, pos=0.5, inner sep=1.0, rotate=90] {\Large ...};
				
				\draw [->, >=latex, dblue] (31, 7) -- (31, 9);
				\draw (27, 9) rectangle ++(6, 4) node [text width=1cm, align=center, pos=0.5, font=\small\linespread{1.0}\selectfont] {rect. weights};
				
				\draw [->, >=latex, dblue, rounded corners=1] (27, 10) -- (15, 10) -- (13.7, 11.3);
				\draw [->, >=latex, dblue] (20, 10) -- (18.7, 11.3);
				\draw [->, >=latex, dblue] (25, 10) -- (23.7, 11.3);
				
				\path [fill=white] (13, 8.6) circle [radius=0.3];
				\path [fill=white] (13, 7.8) circle [radius=0.3];
				\draw [->, >=latex, dblue] (13, 7) -- (13, 11);
				\path [fill=white] (18, 7.6) circle [radius=0.3];
				\path [fill=white] (18, 10) circle [radius=0.3];
				\draw [->, >=latex, dblue] (18, 7) -- (18, 11);
				\path [fill=white] (23, 10) circle [radius=0.3];
				\draw [->, >=latex, dblue] (23, 7) -- (23, 11);
				
				\draw [dblue] (13, 12) circle [radius=1] node {\large $\times$};
				\draw [dblue] (18, 12) circle [radius=1] node {\large $\times$};
				\draw [dblue] (23, 12) circle [radius=1] node {\large $\times$};
				
				\draw [->, >=latex, dblue] (13, 13) -- (13, 15);
				\draw [->, >=latex, dblue] (18, 13) -- (18, 15);
				\draw [->, >=latex, dblue] (23, 13) -- (23, 15);
				
				\draw [dblue] (13, 17) circle [radius=2] node [black] {\Large $\nobjects$};
				\draw [dblue] (18, 17) circle [radius=2] node [black] {\large $\bsize\super{i}$};
				\draw [dblue] (23, 17) circle [radius=2] node [black] {\large $\bdescription\super{i}$};
				
				\draw [thick, densely dotted, black!60!green, rounded corners=1] (10, -10) rectangle (34, 22);
				\path [draw, black!60!green] (13, 20.5) node {\Large \texttt{RECT}};
				
				\draw (36, -4) rectangle ++(4, 4) node [pos=0.5] {\Large $\bobs_{\texttt{1}}$};
				\draw (36, 3) rectangle ++(4, 4) node [pos=0.5] {\Large $\bobs_{\texttt{2}}$};
				\path [draw] (38, 9.5) node [rotate=90] {\Large ...};
				\draw (36, 12) rectangle ++(4, 4) node [pos=0.5] {\Large $\bobs_{\texttt{T}}$};
				
				\draw [->, >=latex, dblue, rounded corners=1] (25, 17) -- (41, 17) -- (41, -1) -- (43, -1);
				\draw [->, >=latex, dblue] (41, 15) -- (43, 15);
				\draw [->, >=latex, dblue] (41, 6) -- (43, 6);
				
				\path [fill=white] (41, 14) circle [radius=0.3];
				\path [fill=white] (41, 5) circle [radius=0.3];
				\draw [->, >=latex, dblue] (40, -2) -- (43, -2);
				\draw [->, >=latex, dblue] (40, 5) -- (43, 5);
				\draw [->, >=latex, dblue] (40, 14) -- (43, 14);
				
				\draw [dblue, rounded corners=1] (43, -4) rectangle ++(6, 4);
				\path [draw, dblue] (46, -2) node {\Large \texttt{FIND}};
				\draw [dblue, rounded corners=1] (43, 3) rectangle ++(6, 4);
				\path [draw, dblue] (46, 5) node {\Large \texttt{FIND}};
				\draw [dblue, rounded corners=1] (43, 12) rectangle ++(6, 4);
				\path [draw, dblue] (46, 14) node {\Large \texttt{FIND}};
				
				\draw [->, >=latex, dblue] (49, -2) -- (52, -2);
				\draw [->, >=latex, dblue] (49, 5) -- (52, 5);
				\draw [->, >=latex, dblue] (49, 14) -- (52, 14);
				
				\draw [dblue] (54, -2) circle [radius=2] node [black] {\large $\bposition\super{i}_{\texttt{1}}$};
				\draw [dblue] (54, 5) circle [radius=2] node [black] {\large $\bposition\super{i}_{\texttt{2}}$};
				\draw [dblue] (54, 14) circle [radius=2] node [black] {\large $\bposition\super{i}_{\texttt{T}}$};
				\draw (52.5, -10) rectangle ++(3, 3) node [pos=0.5] {\Large $\bm{0}$};
				
				\draw [->, >=latex, dblue, rounded corners=1] (54, 0) -- (54, 1) -- (46, 1) -- (46, 3);
				\draw [->, >=latex, dblue, rounded corners=1] (54, 7) -- (54, 8) -- (46, 8) -- (46, 12);
				\path [draw] (46, 9.5) node [fill=white, inner sep=1.0, rotate=90] {\Large ...};
				\draw [->, >=latex, dblue, rounded corners=1] (54, -7) -- (54, -6) -- (46, -6) -- (46, -4);
			\end{tikzpicture} %
}
		\caption{\rect: rectify overlapping object information.}
		\label{fig:arch-AIR-FIND-RECT}
	\end{subfigure}\hfill
	\begin{subfigure}{0.45\textwidth}
		\centering
		\resizebox{\textwidth}{!}{%
\begin{tikzpicture}[scale=0.2]
			\path (-2, -32) grid (57, 8);
			
			\draw (4, 4) rectangle ++(4, 4) node [pos=0.5] {\Large $\bobs_{\texttt{1}}$};
			\draw [->, >=latex, dblue] (6, 4) -- (6, 2);
			
			\draw [dblue, rounded corners=1] (3, -2) rectangle ++(6, 4) node [pos=0.5] {\Large \texttt{AIR}};
			
			\draw [->, >=latex, dblue] (3, 0) -- (2, 0);
			\draw [->, >=latex, dblue] (6, -2) -- (2.2, -5.4);
			\draw [->, >=latex, dblue] (6, -2) -- (6, -5);
			\draw [->, >=latex, dblue] (6, -2) -- (9.8, -5.4);
			\draw [dblue] (0, 0) circle [radius=2] node [black] {\Large $\nobjects$};		
			\draw [dblue] (1, -7) circle [radius=2] node [black] {\large $\bsize\super{i}$};
			\draw [dblue] (6, -7) circle [radius=2] node [black] {\large $\bdescription\super{i}$};
			\draw [dblue] (11, -7) circle [radius=2] node [black] {\large $\bposition\super{i}_{\texttt{1}}$};
			
			\draw [->, >=latex, dblue, rounded corners=1] (11, -5) -- (11, 0) -- (13, 0);
			\path [fill=white] (8.6, -4.4) circle [radius=0.3];
			\draw [->, >=latex, dblue, rounded corners=1] (7.4, -5.6) -- (10, -3) -- (10, 3) -- (57, 3);
			\path (18.5, 3) -- (22.5, 3) node [pos=0.5, fill=white, inner sep=1] {\Large $...$};
			
			\path [fill=white] (16, 3) circle [radius=0.3];
			\path [fill=white] (25, 3) circle [radius=0.3];
			\path [fill=white] (47, 3) circle [radius=0.3];
			\draw [->, >=latex, dblue] (15, 3) -- (15, 2);
			\draw [->, >=latex, dblue] (24, 3) -- (24, 2);
			\draw [->, >=latex, dblue] (46, 3) -- (46, 2);
			
			\draw (14, 4) rectangle ++(4, 4) node [pos=0.5] {\Large $\bobs_{\texttt{2}}$};
			\draw [->, >=latex, dblue] (16, 4) -- (16, 2);
			
			\draw [dblue, rounded corners=1] (13, -2) rectangle ++(6, 4) node [pos=0.5] {\Large \texttt{FIND}};
			
			\draw [->, >=latex, dblue] (16, -2) -- (16, -5);
			\draw [dblue] (16, -7) circle [radius=2] node [black] {\large $\bposition\super{i}_{\texttt{2}}$};
			
			\draw [->, >=latex, dblue, rounded corners=1] (17.4, -5.6) -- (20, -3) -- (20, 0) -- (22, 0);
			\path (20, -3) -- (20, 0) node [pos=0.5, rotate=90, fill=white, inner sep=1] {\Large $...$};
			
			\path [draw] (20.5, 6) node {\Large $...$};
			\path [draw] (19.5, -10) node {\Large $...$};

			\draw (23, 4) rectangle ++(4, 4) node [pos=0.5] {\Large $\bobs_{\texttt{M}}$};
			\draw [->, >=latex, dblue] (25, 4) -- (25, 2);
			
			\draw [dblue, rounded corners=1] (22, -2) rectangle ++(6, 4) node [pos=0.5] {\Large \texttt{FIND}};
			
			\draw [->, >=latex, dblue] (25, -2) -- (25, -5);
			\draw [dblue] (25, -7) circle [radius=2] node [black] {\large $\bposition\super{i}_{\texttt{M}}$};
			
			\draw [->, >=latex, dblue, rounded corners=1] (26.4, -5.6) -- (32, 0) -- (44, 0);

			\draw (44.5, 4) rectangle ++(5, 4) node [pos=0.5] {\Large $\bobs_{\texttt{M+1}}$};
			\draw [->, >=latex, dblue] (47, 4) -- (47, 2);
			
			\draw [dblue, rounded corners=1] (44, -2) rectangle ++(6, 4) node [pos=0.5] {\Large \texttt{FIND}};
			
			\draw [->, >=latex, dblue] (47, -2) -- (47, -5);
			\draw [dblue] (47, -7) circle [radius=2] node [black] {$\hat{\bposition}\super{i}_{\texttt{M+1}}$};
			
			\draw [->, >=latex, dblue, rounded corners=1] (48.4, -5.6) -- (54, 0) -- (57, 0);
			
			\path [draw] (54, 6) node {\Large $...$};
			\path [draw] (54, -10) node {\Large $...$};		
			
			\draw [->, >=latex, dblue] (2.2, -8.6) -- (5, -11);
			\draw [->, >=latex, dblue] (6, -9) -- (6, -11);
			\draw [->, >=latex, dblue] (9.8, -8.6) -- (7, -11);
			\draw [->, >=latex, dblue] (16, -9) -- (16, -11);
			\draw [->, >=latex, dblue] (25, -9) -- (25, -11);
			\draw [->, >=latex, dblue] (47, -9) -- (47, -11);
			
			\draw [->, >=latex, dblue, rounded corners=1] (45, -7) -- (40, -7) -- (40, -23);
			
			\draw [dblue, fill=white, rounded corners=1] (-1, -15) rectangle ++(57, 4) node [pos=0.5] {\Large \texttt{LSTM}};
			\draw [->, >=latex, dblue] (31, -13) -- (34, -13);
			\path [draw, dblue] (19.5, -13) node {\Large $...$};
			\path [draw, dblue] (54, -13) node {\Large $...$};
			
			\draw [->, >=latex, dblue] (25, -15) -- (25, -17);
			\draw [->, >=latex, dblue] (47, -15) -- (47, -17);
			
			\draw [dblue] (25, -19) circle [radius=2] node [black] {\large $\bmotion\super{i}_{\texttt{M}}$};
			\draw [dblue] (47, -19) circle [radius=2] node [black] {$\hat{\bmotion}\super{i}_{\texttt{M+1}}$};
			
			\draw [->, >=latex, dblue] (25, -24.5) -- (27, -24.5);
			\draw [->, >=latex, dblue, rounded corners=1] (25, -21) -- (25, -29.5) -- (27, -29.5);		
			
			\path [fill=white] (25, -23.5) circle [radius=0.3];
			\path [fill=white] (25, -28.5) circle [radius=0.3];
			\draw [dblue] (23, -7) -- (22, -7) -- (22, -11);
			\draw [->, >=latex, dblue, rounded corners=1] (22, -15) -- (22, -28.5) -- (27, -28.5);
			\draw [->, >=latex, dblue, rounded corners=1] (22, -23.5) -- (27, -23.5);

			\draw [dblue, rounded corners=1] (27, -26) rectangle ++(4, 4) node [pos=0.5] {\Large \texttt{TR}$_{\texttt{p}}$};
			\draw [dblue, rounded corners=1] (27, -31) rectangle ++(4, 4) node [pos=0.5] {\Large \texttt{TR}$_{\texttt{m}}$};
			
			\draw [->, >=latex, dblue] (31, -24) -- (33, -24);
			\draw [->, >=latex, dblue] (31, -29) -- (33, -29);		
			
			\draw [dblue] (35, -24) circle [radius=2] node [black] {$\tilde{\bposition}\super{i}_{\texttt{M+1}}$};
			\draw [dblue] (35, -29) circle [radius=2] node [black] {$\tilde{\bmotion}\super{i}_{\texttt{M+1}}$};
			
			\draw [->, >=latex, dblue] (37, -24) -- (39, -24);
			\draw [->, >=latex, dblue] (37, -29) -- (41, -29);		
			
			\draw [dblue] (40, -24) circle [radius=1] node {\Large \texttt{a}};
			\draw [dblue] (42, -29) circle [radius=1] node {\Large \texttt{a}};
			
			\draw [->, >=latex, dblue, rounded corners=1] (45, -19) -- (42, -19) -- (42, -28);
			\path [fill=white] (42, -24) circle [radius=0.3];
			\draw [->, >=latex, dblue] (41, -24) -- (45, -24);
			\draw [->, >=latex, dblue] (43, -29) -- (45, -29);		
			
			\draw [dblue] (47, -24) circle [radius=2] node [black] {$\bposition\super{i}_{\texttt{M+1}}$};
			\draw [dblue] (47, -29) circle [radius=2] node [black] {$\bmotion\super{i}_{\texttt{M+1}}$};
			
			\draw [->, >=latex, dblue] (49, -24) -- (51, -24);
			\draw [->, >=latex, dblue] (49, -29) -- (51, -29);
			\draw [->, >=latex, dblue] (48.4, -25.4) -- (51, -28);
			\path [fill=white] (49.5, -26.5) circle [radius=0.3];
			\draw [->, >=latex, dblue] (48.4, -27.6) -- (51, -25);
			
			\draw [dblue, rounded corners=1] (51, -26) rectangle ++(4, 4) node [pos=0.5] {\Large \texttt{TR}$_{\texttt{p}}$};
			\draw [dblue, rounded corners=1] (51, -31) rectangle ++(4, 4) node [pos=0.5] {\Large \texttt{TR}$_{\texttt{m}}$};
			
			\draw [->, >=latex, dblue] (55, -24) -- (57, -24);
			\draw [->, >=latex, dblue] (55, -29) -- (57, -29);
			
			\draw [thick, densely dotted, magenta, rounded corners=1] (-2, -32) rectangle (57, -4);
			\path [draw, magenta] (54.5, -5.5) node {\Large \texttt{MOT}};	
		\end{tikzpicture} %
}
		\caption{\mot: inferring motion patterns}
		\label{fig:arch-AIR-FIND-MOT}
	\end{subfigure}\quad
	\begin{subfigure}{0.45\textwidth}
		\centering
		\resizebox{\textwidth}{!}{%
\begin{tikzpicture}[scale=0.2]
			\path (-3, -11) grid (56, 22);

			\draw (-3, 11.5) rectangle ++(6, 4) node [pos=0.5] {\Large $\bobs_{\texttt{1:K}}$};
			
			\draw [->, >=latex, dblue] (3, 13.5) -- (6, 13.5);

			\draw [dblue, rounded corners=1] (6, 10.5) rectangle ++(6, 6) node [text width=1cm, align=center, pos=0.5] {\Large \texttt{AIR inf.}};
			
			\draw [->, >=latex, dblue] (12, 13.5) -- (15.7, 17.7);
			\draw [->, >=latex, dblue] (12, 13.5) -- (15, 13.5);
			\draw [->, >=latex, dblue] (12, 13.5) -- (15.7, 9.3);
			
			\draw [dblue] (17.5, 19.5) circle [radius=2.5] node [black] {\large $\hat{\nobjects}_{\texttt{1:K}}$};
			\draw [dblue] (17.5, 13.5) circle [radius=2.5] node [black] {\large $\hat{\bdescription}\super{i}_{\texttt{1:K}}$};
			\draw [dblue] (17.5, 7.5) circle [radius=2.5] node [black] {\large $\hat{\bsize}\super{i}_{\texttt{1:K}}$};
			
			\draw [->, >=latex, dblue] (19.3, 17.7) -- (23, 14);			
			\draw [->, >=latex, dblue] (20, 13.5) -- (23, 13.5);			
			\draw [->, >=latex, dblue] (19.3, 9.3) -- (23, 13);			
			
			\draw [black!60!green, rounded corners=1] (23, 11.5) rectangle ++(7, 4) node [text width=1cm, align=center, pos=0.5] {\Large \texttt{RECT}};
			
			\draw [->, >=latex, dblue] (30, 13.5) -- (33.7, 17.7);
			\draw [->, >=latex, dblue] (30, 13.5) -- (33, 13.5);
			\draw [->, >=latex, dblue] (30, 13.5) -- (33.7, 9.3);
			
			\draw [dblue] (35.5, 19.5) circle [radius=2.5] node [black] {\LARGE $\nobjects$};
			\draw [dblue] (35.5, 13.5) circle [radius=2.5] node [black] {\Large $\bdescription\super{i}$};
			\draw [dblue] (35.5, 7.5) circle [radius=2.5] node [black] {\Large $\bsize\super{i}$};
			
			\draw [->, >=latex, dblue] (38, 13.5) -- (49, 13.5);
			\path [fill=white] (40, 13.5) circle [radius=0.4];
			
			\draw [->, >=latex, dblue] (38, 7) -- (50.45, 0.05);
			\path [fill=white] (40, 5.75) circle [radius=0.4];

			\draw [dblue] (37.3, 11.8) -- (40, 9);
			\draw [dblue] (37.3, 5.8) -- (39, 4);
			\draw [->, >=latex, dblue, rounded corners=1] (37.3, 17.8) -- (40, 15) -- (40, 5) -- (35, 0) -- (12, 0);
			
			\draw (49, 18) rectangle ++(7, 4) node [pos=0.5] {\Large $\bobs_{\texttt{1:T}}$};
			
			\draw [->, >=latex, dblue] (52.5, 18) -- (52.5, 15.5);			

			\draw [black!40!orange, rounded corners=1] (49, 11.5) rectangle ++(7, 4) node [text width=1cm, align=center, pos=0.5] {\Large \texttt{FIND}};
			
			\draw [->, >=latex, dblue] (43, 13.5) -- (51.5, 0);
			
			\draw [->, >=latex, dblue] (52.5, 11.5) -- (52.5, 8.5);
			
			\draw [dblue] (52.5, 6) ellipse [x radius=2.5, y radius=2.5] node [black] {\large $\hat{\bposition}\super{i}_{\texttt{1:T}}$};

			\draw [->, >=latex, dblue] (52.5, 3.5) -- (52.5, 0);
			
			\draw [magenta, rounded corners=1] (47, -4) rectangle ++(9, 4) node [text width=1cm, align=center, pos=0.5] {\Large \texttt{MOT}};

			\draw [->, >=latex, dblue] (47, -2) -- (43, -2);
			\draw [->, >=latex, dblue] (51.5, -4) -- (52.5, -6);
			\draw [->, >=latex, dblue] (51.5, -4) -- (46.7, -6.5);
			\draw [->, >=latex, dblue] (51.5, -4) -- (40, -6.5);
			
			\draw [dblue] (38.5, -8.5) ellipse [x radius=2.5, y radius=2.5] node [black] {\large $\bmotion\super{i}_{\texttt{M:T}}$};
			\draw [dblue] (45, -8.5) ellipse [x radius=3, y radius=2.5] node [black] {\large $\tilde{\bposition}\super{i}_{\texttt{M+1:T}}$};
			\draw [dblue] (52.5, -8.5) ellipse [x radius=3.5, y radius=2.5] node [black] {\large $\tilde{\bmotion}\super{i}_{\texttt{M+1:T}}$};
			
			\draw [dblue] (40.5, -2) ellipse [x radius=2.5, y radius=2.5] node [black] {\large $\bposition\super{i}_{\texttt{1:T}}$};
			
			\draw [->, >=latex, dblue] (38, -2) -- (12, -2);
			
			\draw [dblue, rounded corners=1] (6, -4) rectangle ++(6, 6) node [text width=1cm, align=center, pos=0.5] {\Large \texttt{AIR gen.}};
			
			\draw [->, >=latex, dblue] (6, -1) -- (3, -1);
						
			\draw (-3, -3) rectangle ++(6, 4) node [pos=0.5] {\Large $\hat{\bobs}_{\texttt{1:T}}$};

		\end{tikzpicture} %
}
		\caption{\acs{ours}: \acl{ours}}
		\label{fig:full_architecture}
	\end{subfigure}
	
	\caption{
		The computational flow of the building blocks \find, \rect, and \mot is depicted in (a)--(c).
		Successive figures abstract them to their interface, indicated by colors.
		The full \ac{ours} is depicted in (d).
	}
	\label{fig:architectures}
\end{figure*}

\begin{figure*}
	\centering
	\begin{subfigure}{\textwidth}
		\resizebox{\textwidth}{!}{
			\begin{tikzpicture}[scale=0.2]
			\path (0, 9) grid (60, 20);
			
			\foreach \x in {1,...,20}
			{
				\path [draw, dblue] (6.3+2.62*\x, 19) node {\tiny \texttt{\x}};
			}
			
			\path [draw, dblue] (3.5, 16.75) node {\Large \texttt{AIR}};
			\draw [dblue] (7, 18) -- (7, 15.6);
			
			\node[inner sep=0pt] at (33.8, 16.8) {\includegraphics[width=10.5cm]{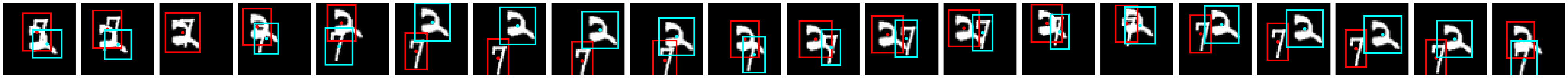}};
			
			\path [draw, dblue] (3.5, 12) node {\large \texttt{VTSSI}};
			\draw [dblue] (7, 14.5) -- (7, 9.5);
			
			\node[inner sep=0pt] at (33.8, 12) {\includegraphics[width=10.5cm]{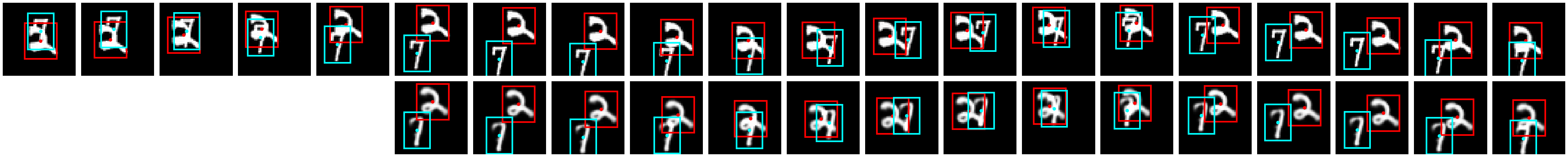}};
			\path [draw, dblue] (12.6, 10.5) node {\texttt{prediction}};
			\draw [->, >=latex, dblue] (17.8, 10.6) -- (19.8, 10.6);
			\end{tikzpicture}
		}
	\end{subfigure}
	\caption{
		Qualitative comparison of \ac{ours} vs.\ \ac{AIR}.
		\ac{AIR} exhibits (i) label switches (frames 5/6, 9/10, 14/15, 19/20), (ii) confusion with overlapping objects (initial frames), resulting in (iii) wrong object count (frame 3).
		\ac{ours} does not exhibit those properties.
		The prediction row shows fully generative samples seeded by inference up until frame 5.
		Other rows show  ground truth frames with $\bposition_t\super{i}$ and $\bsize_t\super{i}$ superimposed.
	}
	\label{fig:qualitative-comparison-ours}
\end{figure*}

\section{Methods}

\subsection{Modifications to \ac{AIR}}\label{sub:modifications}

We use both the generative and the inference model of \ac{AIR} as building blocks of our sequential model.
In particular, we attempt to be faithful to one of its central properties:
the latent space decomposes into a set of distinct objects, each with a set of structured and partially interpretable properties.

In comparison to vanilla \ac{AIR}, we applied two modifications described below (and in more detail in \cref{app:modifications}), leading to increased training stability, as \ac{AIR} is known to be hard to train \parencite{kosiorek_sequential_2018}.
\paragraph{Continuous Counting} 
Defying the low-variance gradient estimates of reparameterized random variables, discrete counting variables are difficult to integrate in VAE-flavored models.
They typically require variance reduction techniques \parencite{DBLP:conf/icml/MnihR16,mnih_neural_2014} or continuous relaxations \parencite{jang_categorical_2017,maddison_concrete_2017}.

Instead of recurrent binary one-step decisions as suggested by \textcite{eslami_attend_2016}, we suggest a feed-forward block that returns a real-valued variable $\cnt$,
which is turned into a sequence of ones equal in length to the integer part of $\cnt$, followed by the remaining fractional part, followed by an appropriate number of zeroes up until the maximum number of objects;
\eg $\cnt= 2.4$ is turned into the sequence $[1, 1, 0.4, 0, \dots]$.
The elements of the list are used to multiply the glimpses $\bglimpse\super{i}$ in the generative part, cf.\ \cref{subfig:continuous-counting}.
In contrast to rounding, this forces the counting variable to take on values close to an integer in order to minimize reconstruction error.
Optimizing on the fractional remainder is inspired by \textcite{graves_adaptive_2016-1}, where this technique regulates the number of computation steps in a recurrent neural network.

\paragraph{Centering Objects in Bounding Boxes}
We found the inference model of AIR to struggle with centering objects within bounding boxes.
In the static case, this is not sufficiently detrimental to reconstruction performance.
However, if the position is not reliably detected in the center of an object, position prediction in our dynamic scenario is difficult. 

We countered this phenomenon with a simple regularization: before being pasted onto the canvas, each glimpse $\bglimpse\super{i}$ is multiplied by a mask of values in $(0, 1]$ that fades out towards the edges like a bell curve, highlighting on the center.
The procedure is depicted in \cref{subfig:centering-bounding-boxes}, as well as exemplary bounding boxes from models trained without and with the regularization.
Over training, all mask values increase monotonically to $1$.

\subsection{Sequential Components}\label{sub:sequential-components}
Applying \ac{AIR} independently to every frame neglects temporal consistency.
Closer analysis reveals three core challenges extending \ac{AIR} to sequential data, exemplified in \cref{fig:qualitative-comparison-ours}.
We discuss each challenge in the subsequent sections and target each with a respective architecture component, culminating in a sequential generative model and inference framework we call \acf{ours}.
Extensive implementation details can be found in \cref{app:implementation}.

\subsubsection{Prevent Label Switching}
\label{sub:find}
The order of attention in \ac{AIR} is arbitrary.
Empirically, it learns a spatial policy for attention order, \eg left-to-right, top-to-bottom \parencite{eslami_attend_2016}.
With moving objects, this inevitably leads to permutations in object discovery order between frames.

The first component aims at preventing label switches.
Rather than independently discovering objects with \ac{AIR} in every frame, we start from an object description $\bdescription\super{i}$ obtained from the first frame and try to find the corresponding object in subsequent frames.
In comparison to \ac{AIR}, this reverses the inference order of object \emph{position} $\bposition\super{i}$ and \emph{description} $\bdescription\super{i}$.
This prevents label switches, while reducing the number of applications of the computationally expensive AIR component from $T$ to one.
The implementation is inspired by the fast-weights approach \parencite{schmidhuber_learning_1992,ba_using_2016}:
we compute convolution kernels from $\bdescription\super{i}$.
From the resulting features of frame $\bobs_{t}$  and the previous position $\bposition\super{i}\tm$  the updated position $\bposition\super{i}_t$ is inferred.
Since its task is to \emph{find} a previously seen object, we call this component \find.
It is depicted schematically in \cref{fig:arch-AIR-FIND}.

\subsubsection{Inference for Overlapping Objects}
\label{sub:rect}
In a single frame, \ac{AIR} cannot distinguish between multiple overlapping objects and non-overlapping regular objects, since it is not equipped with a semantic understanding of the difference between the two or any other prior information as to the appearance of the objects it is supposed to detect.

If we can assume non-overlapping objects in the first frame, \ac{AIR} can provide a concise object description $\bdescription\super{i}$, and the \find module will maintain consistent object order throughout the sequence.
We introduce the second component \rect (for rectification) to relax this assumption: rather than relying on \ac{AIR}'s object description from the first frame, a \ac{RNN} processes the inference output on the first $K$ frames.
This net reaches a consensus $\bstate\super{i}$ from the $K$ sets $\hat{\bstate}\tsub{1}{K}\super{i}$ of latent variables from applications of \ac{AIR} on the first $K$ frames, \eg by means of weighted averaging.
Finally, we use the more robust consensus $\bdescription\super{i}\in\bstate\super{i}$ as the input to the \find module.
This procedure is depicted in \cref{fig:arch-AIR-FIND-RECT}.

\subsubsection{State-space Modeling of Motion}

\label{sub:mot}
Operating on individual frames, \ac{AIR} cannot incorporate the governing motion law, hence fails to predict likely future paths from an object's history.

\find and \rect are designed to deal with label switches and object overlap.
This is largely achieved by improving the inference of object positions across time compared to vanilla \ac{AIR}.
The third component introduces a dynamical system to the position variable.
In contrast to \find and \rect, this affects both the generative and the inference model:
the state-space model (SSM) assumption requires us to add an explicit motion random variable $\bmotion\super{i}_t$ to the latent space.
It captures higher-order motion description, \eg velocities, accelerations, or curve radii.
This allows us to define Markov transition priors $\p*{\bposition_t\super{i},\bmotion_t\super{i}}{\bposition\tm\super{i},\bmotion\tm\super{i}}$
for state prediction in the next frame given the current state.
As we will show empirically---cf.\ \cref{sec:experiments}---this allows faithful multi-step object-level prediction. 
To infer the motion variable, we feed the object position proposals $\hat{\bposition}\Ts$ from \find to an \ac{RNN}.
After $M$ frames, where $M$ is at least the order of dynamics assumed, the \ac{RNN} provides inferred motion proposals $\hat{\bmotion}\tsub{M\shortplus1}{T}$.
Both position and motion proposals are fused with prior predictions $\tilde{\bposition}_t\super{i}$ and $\tilde{\bmotion}_t\super{i}$ from the transition prior.
The fusion is achieved by averaging.
The procedure is depicted in \cref{fig:arch-AIR-FIND-MOT}.

\subsection{\acl{ours}}\label{sub:ours}

Combining all suggested modules, we arrive at the full architecture, which we call \acf{ours}.
It processes initial frames $\khead$ separately with \ac{AIR}; 
reaches a consensus with \rect; uses this consensus in \find to determine positions; 
refines the position estimates with dynamic information by exploiting \mot.
This procedure is depicted in \cref{fig:full_architecture}.
For the generative model, the major change towards \ac{AIR} is the Markovian evolution of positions $\bposition_t$ and motion descriptions $\bmotion_t$ over time.

The model is trained with stochastic gradient descent on the sequential evidence lower bound (ELBO)
\begin{align*}
	&\;\expc[
		\q%
	]{
		\ln\frac{
			\p*{\bobs\Ts,\nobjects,\set{\bposition\Ts\super{i},\bmotion\tsub{M}{T}\super{i},\bdescription\super{i},\bsize\super{i}}}
		}{
			\q*{\nobjects,\set{\bposition\Ts\super{i},\bmotion\tsub{M}{T}\super{i},\bdescription\super{i},\bsize\super{i}}}{\bobs\Ts}
		}
	}\\&\;\leq\ln\p{\bobs\Ts}.
\end{align*}
Factorizations for $p$ and $q$ can be found in \cref{app:math},
implementation details in \cref{app:implementation}.

An interesting feature of \ac{ours} is its modularity:
rather than using the full model with all suggested components, we can choose to use only some of them depending on the downstream task for a more efficient model.
We will investigate this in the following section.

\subsection{Evaluating Components of \ac{ours}}
\label{sub:ablation}
\renewcommand\theadfont{\tiny}

\newcommand{\tabheader}{\makecell[r]{count\\acc.} & \makecell[r]{inf.\\error} &\makecell[r]{pred.\\error}}
\newcommand{\shorttabheader}{\tabheader}
\begin{table*}
	\caption{
		Quantitative tracking and prediction results with variants of \acf{ours} on 10000 test set trajectories.
		Counting accuracy refers to the average percentage of frames for which the amount of present objects is determined correctly.
		Inference and prediction errors refer to the average per-frame Euclidean distance (unit: pixels) from the inferred or predicted object center to the ground truth, respectively.
	}
	\label{tab:results}
	\centering
	\resizebox{\textwidth}{!}{
		\begin{tabular}{@{}cc *{3}{crrr} *{2}{crrr}@{}}
			\toprule
			&&& \multicolumn{3}{c}{\ac{AIR}} && \multicolumn{3}{c}{\find} && \multicolumn{3}{c}{\rect/\find} && \multicolumn{3}{c}{\find/\mot} && \multicolumn{3}{c}{\ac{ours}} \\
			\cmidrule{4-6} \cmidrule{8-10} \cmidrule{12-14} \cmidrule{16-18} \cmidrule{20-22}
			motion&\makecell{overlap in\\1st frame}&& \shorttabheader && \shorttabheader && \shorttabheader && \tabheader && \tabheader\\ \cmidrule{1-2} \cmidrule{4-22}
			linear   & \xmark && 97.63\% & 5.953 &n/a&& 99.98\% & 1.019 &n/a&& 99.99\% & 1.131 &n/a&& 99.97\% & 1.291 & 3.491 && 99.99\% & 1.035 & 3.442\\
			& \cmark && 97.22\% & 5.620 &n/a&& 91.53\% & 2.739 &n/a&& 99.70\% & 1.225 &n/a&& 92.67\% & 3.002 & 5.197 && 99.50\% & 1.109 & 3.544\\
			elliptic & \xmark && 97.33\% & 5.160 &n/a&& 99.98\% & 0.973 &n/a&& 99.98\% & 1.095 &n/a&& 99.99\% & 0.846 & 2.836 && 99.96\% & 1.028 & 2.583\\
			& \cmark && 96.72\% & 4.833 &n/a&& 90.95\% & 2.130 &n/a&& 99.48\% & 1.194 &n/a&& 89.55\% & 2.282 & 4.365 && 99.54\% & 1.076 & 2.676\\
			\bottomrule
		\end{tabular}
	}
\end{table*}

We study five models corresponding to the architectures depicted in \cref{fig:air-architecture,fig:architectures}:
\begin{enumerate}
\item \ac{AIR} (with modifications from \cref{sub:modifications}), 
\item \find (based on AIR), 
\item \rect/\find (\ie \ac{ours} without \mot), 
\item \find/\mot (\ie \ac{ours} without \rect) and 
\item full \ac{ours}.
\end{enumerate}

We trained these variants on four flavors of Moving MNIST---we use several variants with different features to perform targeted studies of the components of \ac{ours}: 
the data show either linear or elliptic motion, and either the first frame is guaranteed to contain only non-overlapping digits or not.
We evaluated object counting accuracy as a proxy for robustness towards overlapping digits.
Further, we report the accuracy of the position inference against ground truth, as well as prediction accuracy for the two models that make use of \mot (all other models cannot generate coherent sequences by design).
The results can be found in \cref{tab:results}.
We make several interesting observations:

\find drastically improves the inference accuracy when the first frame is sufficiently clean to identify objects. 
In fact, \find is on a par with \ac{ours} in these scenarios, despite being much more lightweight.
\ac{AIR} suffers from label switches and recounting every frame.
The results for \find drop significantly when the assumption of non-overlapping objects in the first frame is removed.
This can be mitigated by the introduction of \rect.
We hypothesize that the slight drop in performance compared to \find on non-overlapping first frames hints at room for improvement with the consensus mechanism of \rect.

\begin{figure*}
	\resizebox{\textwidth}{!}{
		\begin{tikzpicture}[scale=0.2]
		\path (0, -2) grid (60, 8);
		
		\path [draw, dblue] (3.5, 4.8) node {\texttt{FIND}};
		\path [draw, dblue] (3.5, 1.8) node {\texttt{VTSSI}};
		\path [draw, dblue] (3.5, -1.2) node {\texttt{SQAIR}};
		
		\foreach \x in {1,...,10}
		{
			\path [draw, dblue] (6.3+2.5*\x, 7) node {\tiny \texttt{\x}};
		}
		
		\foreach \x in {1,...,10}
		{
			\FPeval{\fid}{clip(\x+5)}
			\path [draw, dblue] (33.9+2.5*\x, 7) node {\tiny \texttt{\fid}};
		}
		
		\node[inner sep=0pt] at (20, 4.8) {\includegraphics[width=5cm]{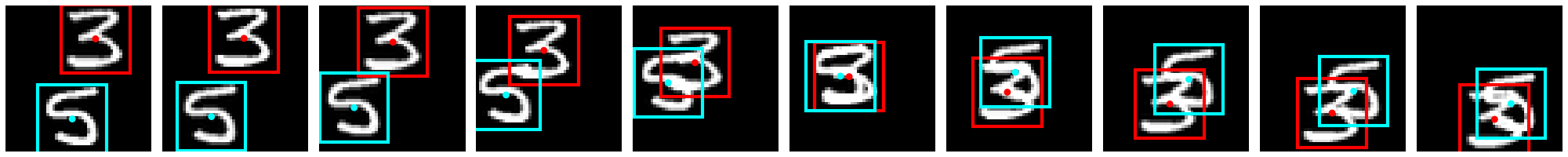}};
		\node[inner sep=0pt] at (47.6, 4.8) {\includegraphics[width=5cm]{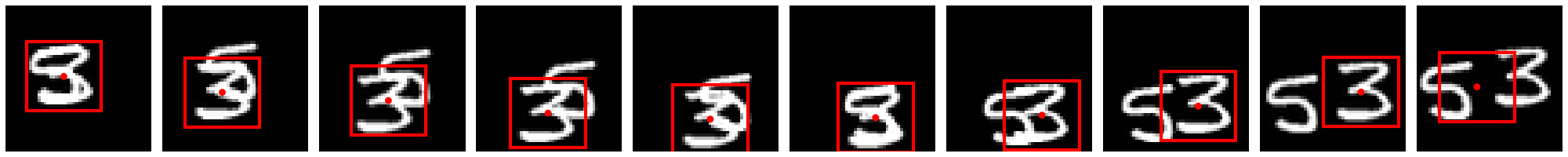}};
		
		\node[inner sep=0pt] at (20, 1.8) {\includegraphics[width=5cm]{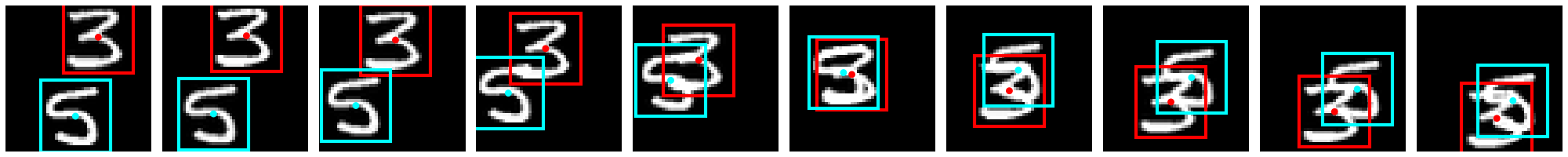}};
		\node[inner sep=0pt] at (47.6, 1.8) {\includegraphics[width=5cm]{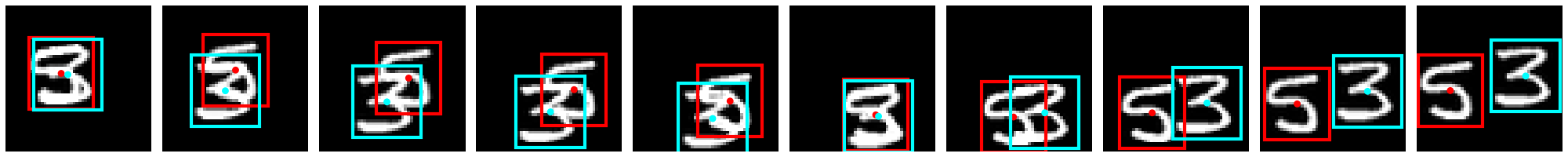}};
		
		\node[inner sep=0pt] at (20, -1.2) {\includegraphics[width=5cm]{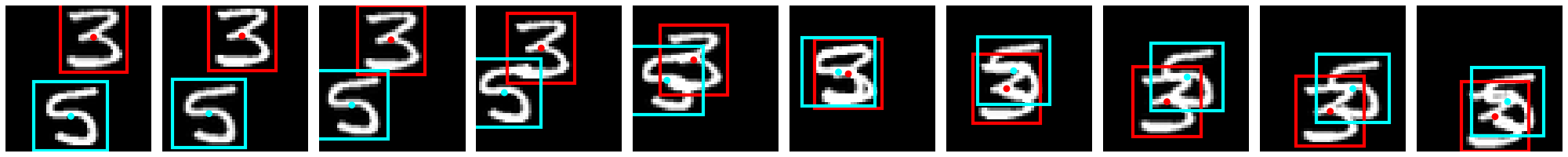}};
		\node[inner sep=0pt] at (47.6, -1.2) {\includegraphics[width=5cm]{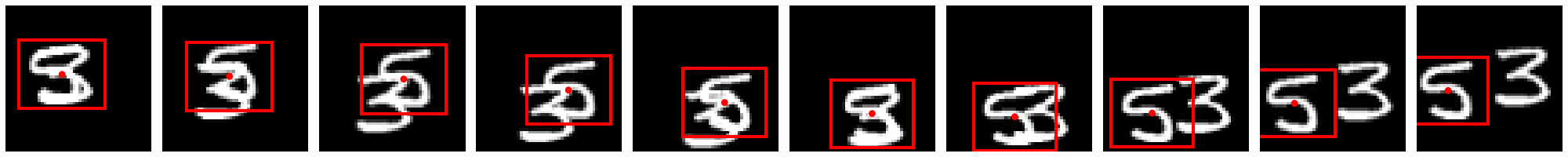}};
		
		\draw [dblue] (33.8, 6) -- (33.8, -2.5);
		
		\end{tikzpicture}
	}
	\caption{
		Qualitative example of challenging overlap.
		In the left half, \find, \ac{ours}, and SQAIR successfully infer object properties.
		The right half shows the same sequence, but the first five frames were dropped so that the initial frame is cluttered.
		\find and SQAIR, relying on \ac{AIR} for discovery, only recognize one object, and are unable to correct.
		\ac{ours} recognizes both digits, despite overlap in all $K=5$ first frames.
	}
	\label{fig:qualitative-prediction}
\end{figure*}
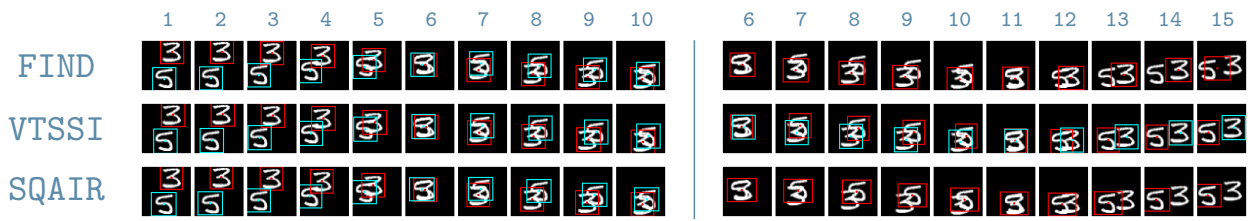

\rect is very robust \wrt overlapping objects, as 
\cref{fig:qualitative-prediction} highlights.
\find, SQAIR \parencite{kosiorek_sequential_2018}, and the \rect-based \ac{ours} successfully tackle the sequence on the left side with a clean first frame.
When these models do not get access to the first five frames, but start with the cluttered frames 6 and higher, \find and SQAIR are unable to recover from the wrong count in the first frame.
This is a consequence of \ac{AIR}'s inability to deal with overlapping frames.
We note that \ac{ours} succeeds despite \rect only accessing $K=5$ frames (\ie frames 6--10 in this case), all of which have overlapping objects.

\mot by itself generally does not lead to improved \emph{inference} over \find.
When combined with \rect to form \ac{ours}, however, generative accuracy increases, even for scenarios where \rect is not strictly necessary---being able to predict helps inference.
The full \ac{ours} handles all variants equally well.
It performs well on linear and non-linear motion, with slight advantage on the non-linear, but smooth elliptic movements compared to discontinuous bouncing behavior, which is more difficult to predict.

We conclude that each component fulfills its designated purpose:
in the absence of \find, we observe label switching;
in the absence of \rect, overlapping objects cannot be disentangled reliably;
in the absence of \mot, prediction is impossible, but even inference performance drops slightly.

Our evaluation also suggests that we can take advantage of the modular composition of \ac{ours}.
For inference, \find and \rect are the decisive factors.
If prediction is not necessary, we can reliably train and use a simpler model. %

\section{Related Work}\label{sec:related_work}
Multi-object tracking has been the primal concern of many works \parencite{pulford_taxonomy_2005}.
\textcite{DBLP:conf/icip/BewleyGORU16} propose using a detector and a subsequent state-space model, showing the promise of such methods outside a deep learning context.
\textcite{DBLP:journals/corr/abs-1210-3288} formulate tracking as a mixture of Dirichlet processes operating on top of a feature extraction pipeline without the need for supervision signals.
A series of works considers tracking via end-to-end supervised learning \parencite{DBLP:conf/cvpr/KahouMMPV17,DBLP:conf/nips/KosiorekBP17,DBLP:journals/ral/GordonFF18,DBLP:conf/iscas/NingZHRWCH17}, showing that it is possible to represent trackers with neural architectures when annotated data is available.

In video prediction the central concern is the prediction of future frames in a video stream \parencite{DBLP:conf/icml/SrivastavaMS15,DBLP:conf/iclr/BabaeizadehFECL18,DBLP:conf/icml/DentonF18,DBLP:journals/corr/abs-1804-01523}.
This can be expressed as inference in the underlying generative model, but without a focus on tracking.
This is the starting point of our method, which is based on variational sequence models \parencite{bayer_learning_2014,chung_recurrent_2015-1}.
We rely on a state-space formulation where the graphical model has Markov properties \parencites{DBLP:books/daglib/0039111}; this has been pioneered in a neural variational context by \textcite{DBLP:journals/corr/KrishnanSS15,archer2015black,DBLP:conf/nips/FraccaroSPW16,DBLP:conf/iclr/KarlSBS17}.
Note that \textcite{DBLP:conf/iclr/SteenkisteCGS18} also perform scene decomposition with neural networks.

\subsection{Relation to DDPAE and SQAIR}
\label{sub:sqair-ddpae}
Two related approaches to ours have been suggested in the literature: Decompositional Disentangled Predictive Auto-Encoders (DDPAE; \cite{hsieh_learning_2018}) and Sequential \ac{AIR} (SQAIR; \cite{kosiorek_sequential_2018}).
Both approaches use attention-based amortized inference to decompose video sequences of moving objects into per-object latent state sequences.
Like \ac{ours}, both approaches borrow the likelihood model $\p(\bobs_t\mid\{\bstate_t\super{i}\})$ of \ac{AIR}, cf.\ \cref{sub:air}.

DDPAE focuses on faithful prediction of the tail $\ktail$ of a sequence from its head $\khead$.
As a consequence, it is trained on a lower bound to the conditional $\p{\ktail}{\khead}$ rather than the joint $\p{\bobs\Ts}$.
This also leads to architectural differences:
in contrast to \ac{ours} and SQAIR, DDPAE does not auto-encode the entire sequence, but follows a seq2seq-inspired approach \parencite{sutskever_sequence_2014}.
The sequence head $\khead$ is only used for inference and never reconstructed.
Conversely, the latent states $\bstate\tsub{K\shortplus1}{T}$ of the sequence tail $\ktail$ are never inferred from data, but predicted from the head.
Both inference and prediction are implemented by \acp{RNN}.
DDPAE further models interactions between objects by means of another recurrence that connects inference of individual objects.

SQAIR introduces two inference components: PROP and DISC.
PROP handles object propagation between frames.
Two recurrent cells update the position, then (based on the new position) update description and presence.
DISC discovers new objects.
It works much akin to inference in \ac{AIR}, except that the inference of a new object is informed by the latent states of existing objects from propagation to avoid duplicate discovery.
Relying on \ac{AIR} to this extent, SQAIR inherits its inability to handle overlapping objects in inference for the first time step and assumes non-overlapping first frame.
SQAIR can, in principle, support entering and exiting object at arbitrary frames.

Contrasting DDPAE and SQAIR with \ac{ours}, we conclude that all models share the same ancestor \ac{AIR}, specifically the non-dynamic part of latent space design and resulting likelihood model.
A major distinctive feature of this work is enhancing the state space with an explicit motion variable $\bmotion$, capturing the dynamics of motion.
This extra variable turns the position transition fully Markov and the overall model into a proper state-space model.
In contrast, both DDPAE and SQAIR use recurrent cell states in the transition model, which need to capture the motion information.
This reduces the interpretability of the latent state, as the role of the recurrent state is unclear for each specific model, and rules out regularization via priors.

All three models implement significantly different inference procedures for the sequential case.
Our modular framework focuses on robust inference even in challenging scenarios, to allow for accurate long-term prediction, even for complicated non-linear motion.
Object interaction (as in DDPAE) or entering and exiting objects (as in SQAIR) are not considered, but could be introduced by adding new or modified components to the \ac{ours} framework.
\section{Experiments}\label{sec:experiments}
On top of our ablation studies in \cref{sub:ablation}, we study \ac{ours} against the baselines DDPAE and SQAIR.
The experiments investigate the robustness in inference/tracking and prediction, particularly over longer horizons.
We build upon of the Moving MNIST data sets previously studied with the baselines.

All models are trained with stochastic gradient descent on the evidence lower bound.
We used the Adam optimizer \parencite{kingma_adam_2015}.
We borrow the curriculum schedule from SQAIR, where the length of the training sequence is increased over training time.
For details on the training procedure, the experimental setup, and additional results, see \cref{app:curriculum,app:experiments,app:results}.

\subsection{Prediction}\label{sub:exp-prediction}
\begin{figure}
	\centering
	\includegraphics[width=.8\linewidth]{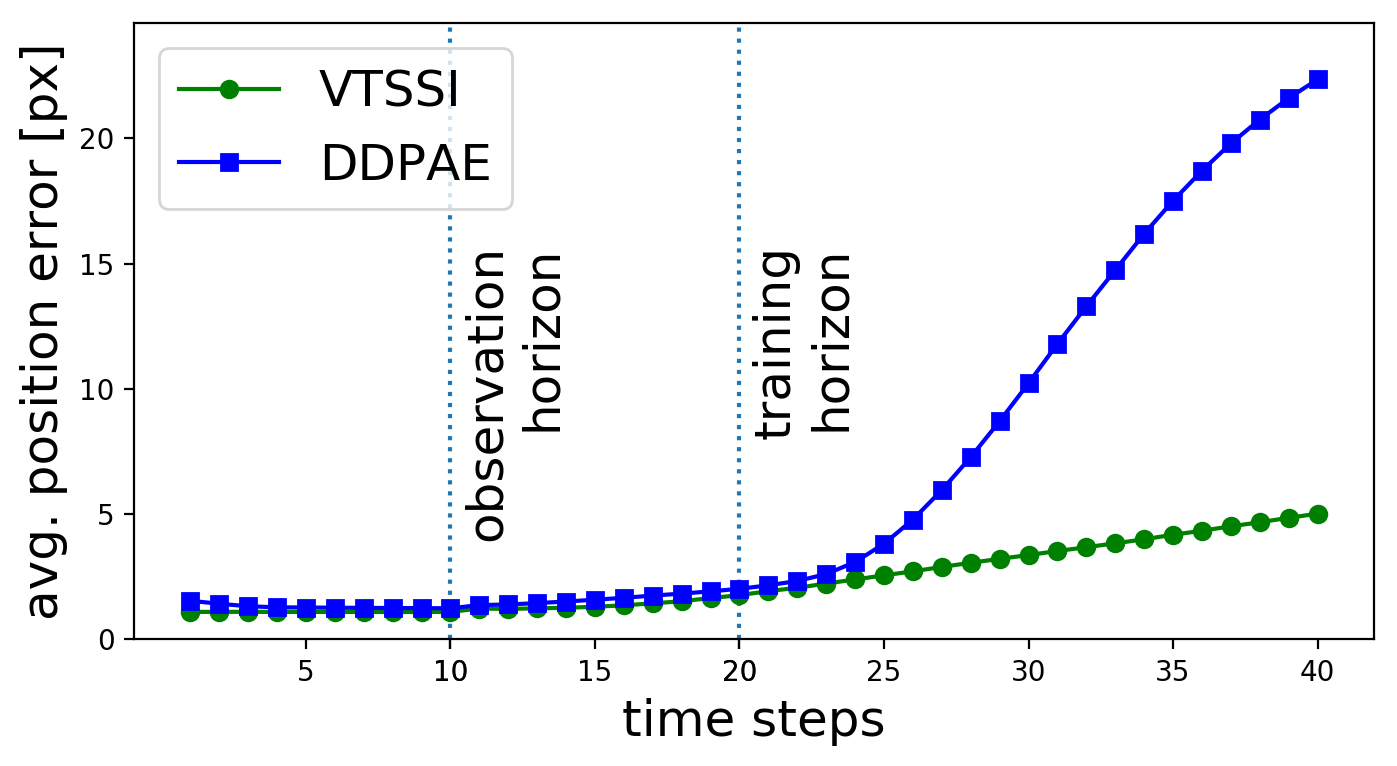}
	\caption{
		Test set prediction errors of DDPAE vs. \ac{ours} on data used in the original publication.
		Details in \cref{sub:exp-prediction}.
	}
	\label{fig:quantitative-ddpae}
\end{figure}
\begin{figure}
	\centering
	
	\begin{subfigure}{.4\textwidth}
		\centering
		\includegraphics[width=\textwidth]{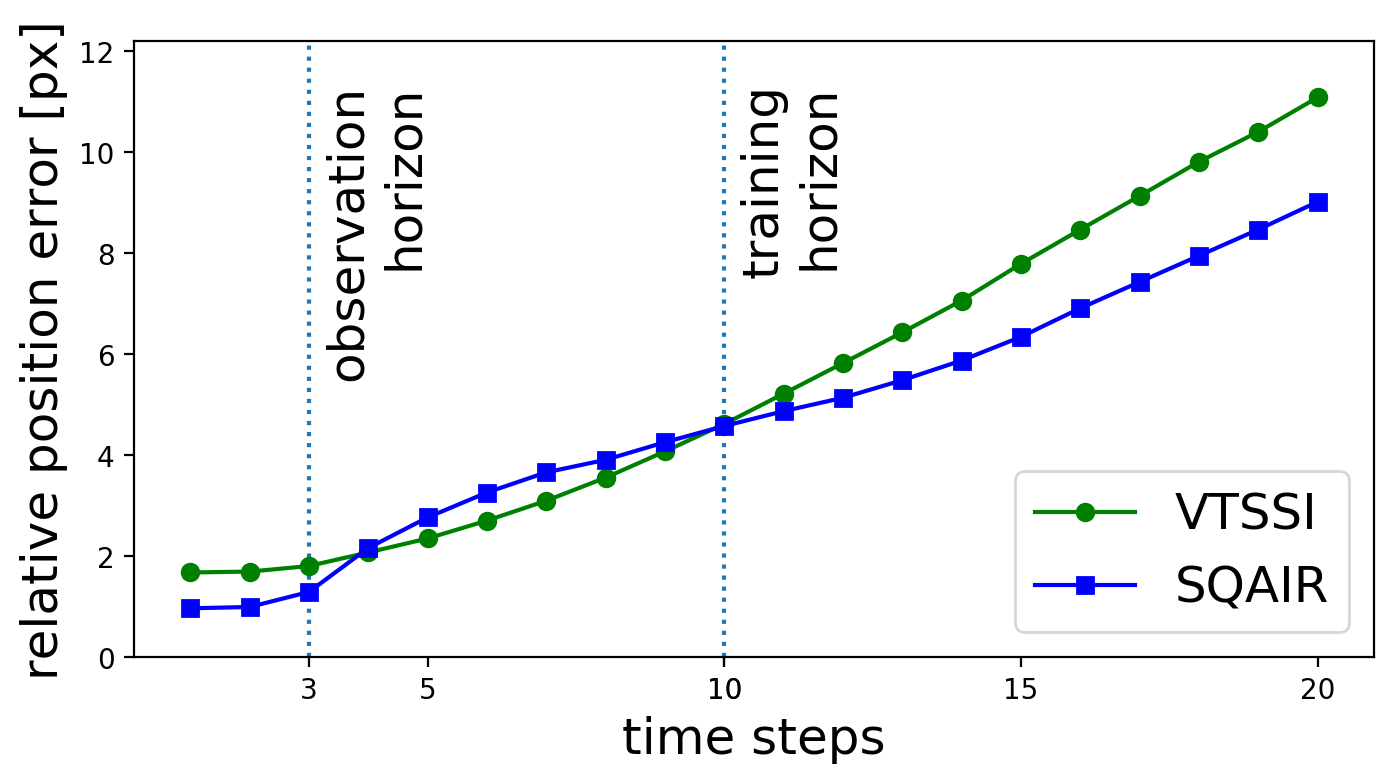}
		\caption{\ac{ours} vs.\ SQAIR: SQAIR data.}
		\label{fig:sqair-prediction-sqair}
	\end{subfigure}
	\quad
	\begin{subfigure}{.4\textwidth}
		\centering
		\includegraphics[width=\textwidth]{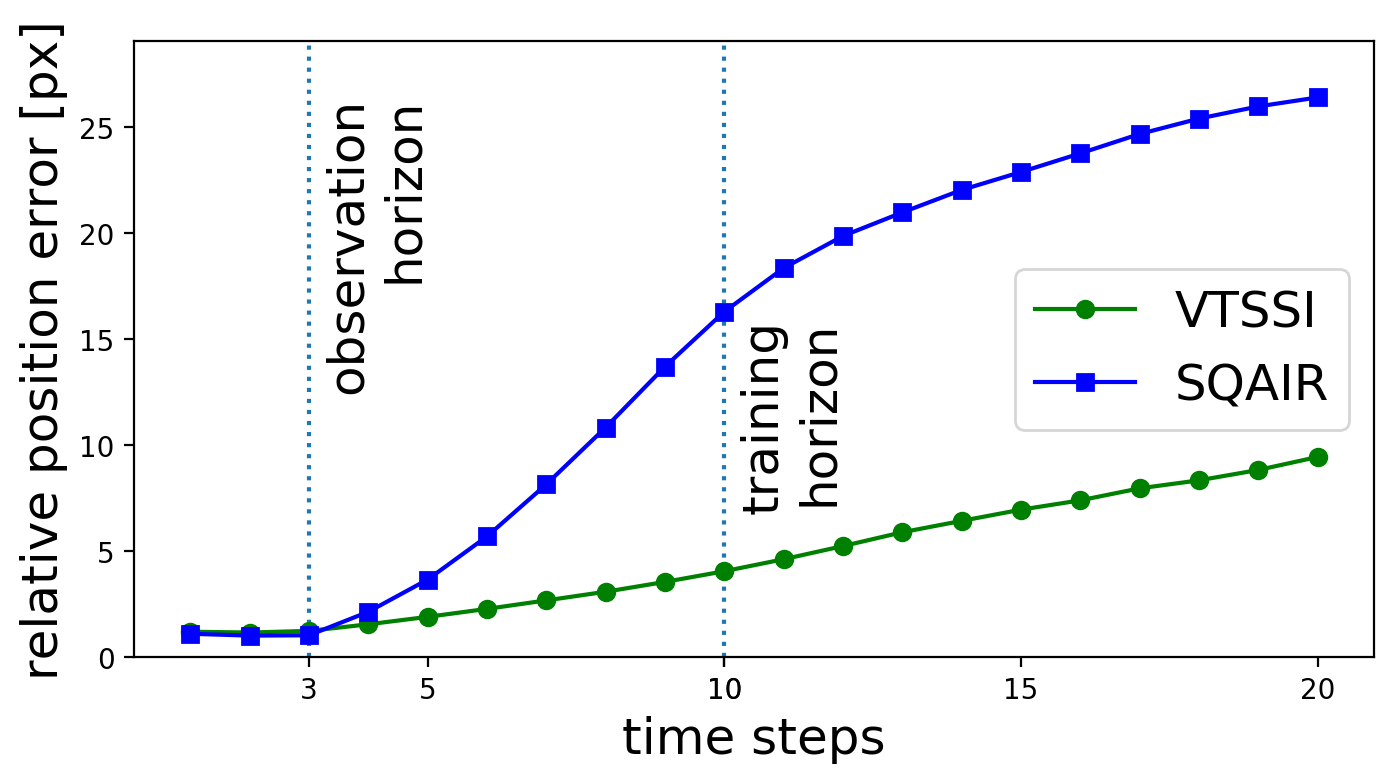}
		\caption{\ac{ours} vs.\ SQAIR: our linear data.}
		\label{fig:sqair-prediction-lin}
	\end{subfigure}

	\caption{
		Test set prediction of SQAIR vs.\ \ac{ours} on its data set and our data set.
				The models perform inference on three observations (first vertical line), the observation horizon. 
				After that, object trajectories are sampled generatively without access to further observations and beyond training sequence length (second vertical line).
	}
	\label{fig:quantitative-prediction}
\end{figure}

\textcite{hsieh_learning_2018} provide a data generation process for DDPAE.
We build a training and test set from this process to ensure fair comparability, cf. \cref{appsub:data}.
We train both models with $T=20$ and $K=M=10$, as in the original publication.
Starting from inferences with $K=10$, we tested the position prediction error for $T>20$, probing the generalization of the learned predictions.
The average performance across a test set of 10000 sequences can be seen in \cref{fig:quantitative-ddpae}.
DDPAE and \ac{ours} are equally faithful to ground truth within the training horizon.
However, the recurrent prediction cell of DDPAE is unable to generalize beyond the training horizon, it seems to severely overfit on the training horizon.
This is particularly remarkable given DDPAE's loss is tailored towards prediction.

As with DDPAE, we tried to compare VTSSI to SQAIR on its original data set.
\textcite{kosiorek_sequential_2018} also provide a data generation process.
We used the same data generation process, except we removed the noise, which turned prediction comparisons in the confined frames futile.
We train both models with $T=10$ and $K=M=3$, as in the original publication.
On these data, we find SQAIR and \ac{ours} to perform equally well, with slight advantage for \ac{ours} within the training horizon, and for SQAIR outside the training horizon, cf. \cref{fig:sqair-prediction-sqair}.
We noticed a subtle, but crucial difference in the generation process of these data against the data we used for the results in \eg \cref{tab:results}:
the data generation implements bouncing of the walls in terms of the top left corner of the tight bounding box (\ie an object bounces in-frame on the top and left border and out-of-frame otherwise).

To examine the effect, we trained SQAIR on the linear data set suggested in \cref{sub:ablation} with clean first frames, \ie not SQAIR's original data set, but well within SQAIR's assumptions.
This data  does \emph{not} generate bouncing behavior in terms of bounding boxes, but the actual object appearance.
The result can be seen in \cref{fig:sqair-prediction-lin}.
When required to model bouncing behavior, SQAIR falls short of \ac{ours}.
An example highlighting this observation can be found in \cref{fig:sqair-prediction} in  \cref{app:results}.
SQAIR defines object positions in terms of bounding box corners, not the center (as DDPAE and \ac{ours} do).
We believe that this generally makes it harder to learn accurate object dynamics except when the data set reflects this model assumption.
This may lead to instabilities in the recurrent motion propagation cell of SQAIR.
Using the object center makes it easier to use a simpler Markov transition.
Specific motion behavior of an object can be saved into the motion variable $\bmotion$.

\subsection{Inference and Tracking}\label{sub:exp-inference}

\begin{figure}
	\centering
	\begin{subfigure}{.4\textwidth}
		\centering
		\includegraphics[width=\textwidth]{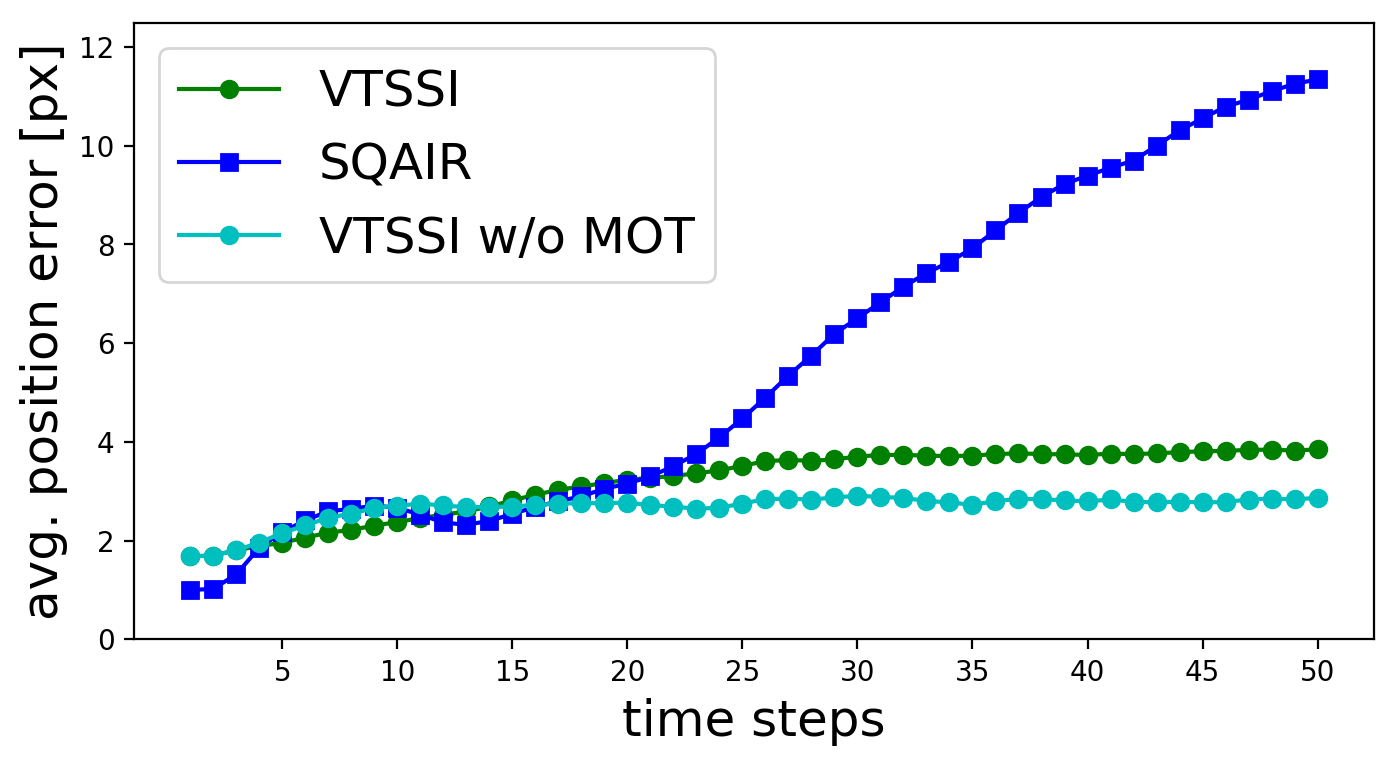}
		\caption{\ac{ours} vs.\ SQAIR: SQAIR data.}
		\label{fig:sqair-inference-sqair}
	\end{subfigure}
	\quad
	\begin{subfigure}{.4\textwidth}
		\centering
		\includegraphics[width=\textwidth]{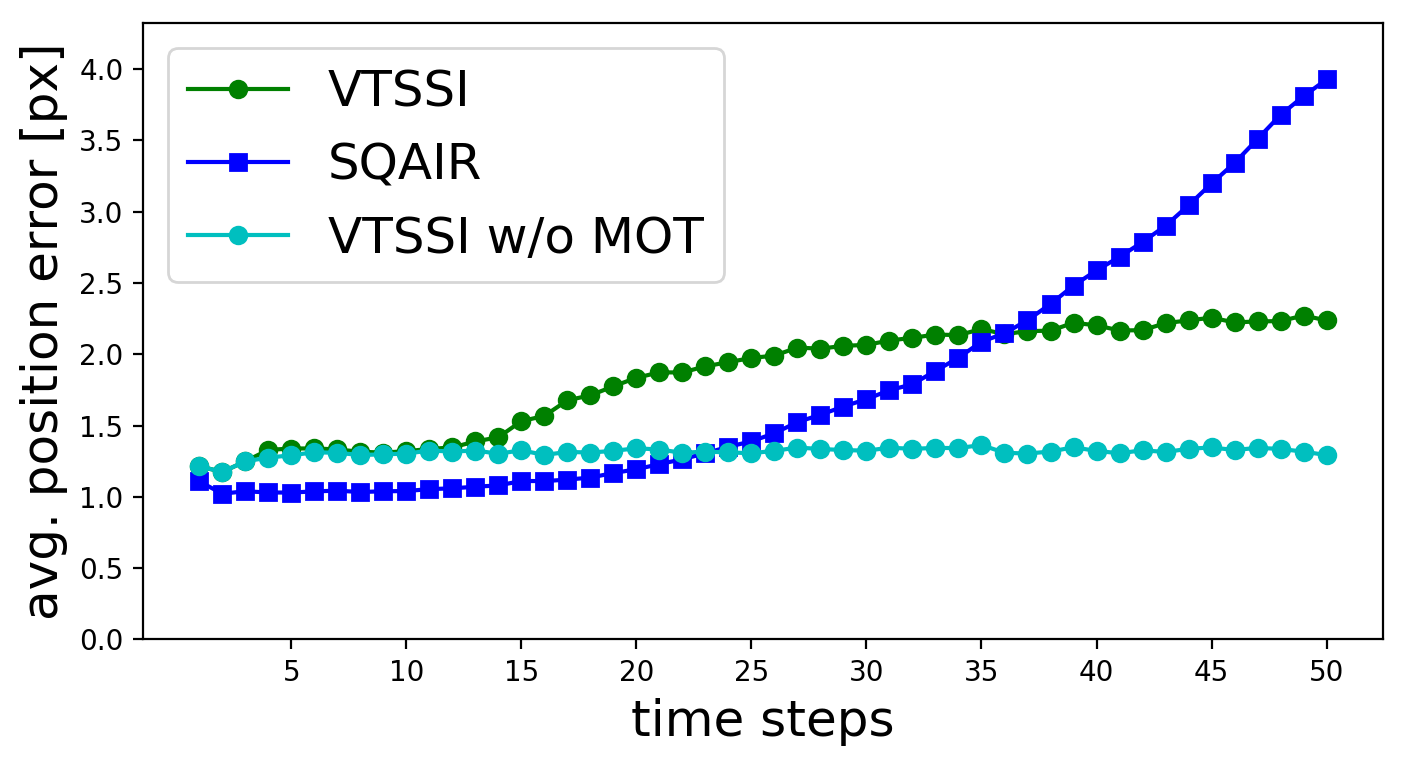}
		\caption{\ac{ours} vs.\ SQAIR: our linear data.}
		\label{fig:sqair-inference-lin}
	\end{subfigure}

	\caption{
		Test set inference error of SQAIR vs.\ \ac{ours} on  
		a noise-free version of its own data set, and our data set.
		Details in \cref{sub:exp-inference}.
	}
	\label{fig:quantitative-inference}
\end{figure}

With the same models and data as in our evaluation of prediction, we also examined tracking performance of SQAIR and \ac{ours}.
The results can be seen in \cref{fig:quantitative-inference}.
On both data sets, we see that the tracking performance of SQAIR drops drastically after around 20 steps.
In contrast, \ac{ours} keeps a constant error over long horizons.

We also added one of the models discussed in \cref{sub:ablation}, \ac{ours} without the \mot component, which is not necessary for pure tracking.
Rather than training a new, reduced model separately, this model is achieved by using the full \ac{ours} model.
At test time, the outputs of its \find component are directly evaluated.
Unaffected by prediction errors, this model achieves even more reliable tracking performance.

An example highlighting this observation can be found in \cref{fig:sqair-tracking} in \cref{app:results}.

\subsection{Discussion}
In the previous analysis, we found that \ac{ours} performs much more robustly in both prediction and inference, especially over long horizons.
We speculate that this can be attributed to the relative simplicity of our model:
where SQAIR and DDPAE use recurrent cells in a black-box fashion, particularly in motion prediction, we use a state-space model in feed-forward fashion with explicit representation of the dynamic state of the object.

Further, our inference model, specifically the \find component, is also of a feed-forward nature (with adaptive convolution kernels).
We believe that this leads to more stable model components, even for long horizons.
Moreover, it drastically reduces the number of applications of the AIR component, particularly compared to SQAIR, which we found to be very beneficial to the robustness.
As a side effect, VTSSI trains significantly faster than SQAIR.
Using reference implementations of the original authors, our model required at least an order of magnitude less wall clock time until convergence.
The amount of parameters was roughly equal and most were used by the AIR base model.

DDPAE and SQAIR each provide orthogonal features not covered by \ac{ours}---object interaction and vanishing objects, respectively.
The flexible modular nature of \ac{ours} allows to add suitable model components for these purposes.
In this work, we chose not to focus on such scenarios, as the presumably simpler scenarios we presented already proved challenging for related models.
We plan to add these features in future work.
\section{Conclusion}\label{sec:discussion}
We introduced \acf{ours},
a generative disentangled state-space model inspired by \acf{AIR} with a modular neural inference procedure.
\ac{ours} successfully decomposes sequences by describing the objects that make up the scene, and learns a state-space model of the observations that is able to predict faithfully over long horizons.
Our experiments show that the inference components that define \ac{ours} form a modular framework that may be tailored to the task at hand.
We further showed that our inference model can overcome limiting assumptions of \ac{AIR}.
In comparison to related state-of-the-art baselines, we significantly improved performance in prediction and tracking. 
\subsubsection*{References}
\printbibliography[heading=none]
\clearpage
\appendix
\makeatletter
\renewcommand{\appendixpagename}{Appendix to\\\emph{\@title}}
\makeatother
\appendixpage

\section{Generative and Inference Model}\label{app:math}
The full generative process is
\begin{align*}
& \p{\bobs\Ts,\nobjects,\set{\bposition\Ts\super{i},\bmotion\tsub{M}{T}\super{i},\bdescription\super{i},\bsize\super{i}}_{i=1}^\nobjects}\\[-1mm]
= \,& \p*{\nobjects}\prod_{i=1}^{\nobjects}\p*{\bdescription\super{i}}\p*{\bsize\super{i}}\prod_{t=1}^{T}\p*{\bobs_t}{\set{\bposition_t\super{i},\bdescription\super{i},\bsize\super{i}}}\\[-1mm]
&\cdot
\p*{\bposition_1\super{i}}\prod_{t=2}^{M}\p*{\bposition_t\super{i}}{\bposition\tm\super{i}}\,\p*{\bmotion_M\super{i}}\\
&\cdot
\prod_{t=M\shortplus1}^{T}\p*{\bposition_t\super{i},\bmotion_t\super{i}}{\bposition\tm\super{i},\bmotion\tm\super{i}}. 
\end{align*}
The inference procedure can be described as
\begin{align*}
&\q{\nobjects,\set{\bposition\Ts\super{i},\bmotion\tsub{M}{T}\super{i},\bdescription\super{i},\bsize\super{i}}}{\bobs\Ts}\\[-1mm]
=\,&
\q[\textrm{RECT}]{\nobjects,\bsize\super{i},\bdescription\super{i}}{\hat{\nobjects}\tsub{1}{K},\hat{\bsize}\super{i}\tsub{1}{K},\hat{\bdescription}\super{i}\tsub{1}{K}}\\
&\cdot\prod_{k=1}^{K}\q[\textrm{AIR}]{\hat{\nobjects}\tsub{k},\hat{\bsize}\super{i}\tsub{k},\hat{\bdescription}\super{i}\tsub{k}}{\bobs_k}\\[-1.5mm] 
&\cdot\prod_{t=1}^{T}\q[\textrm{FIND}]{\hat{\bposition}_t\super{i}}{\bposition\tm\super{i}, \bobs_t, \bdescription\super{i}}\\
&\cdot\prod_{t=1}^{M}\q{\bposition_t\super{i}}{\hat{\bposition}_t\super{i}}\q{\bmotion_M\super{i}}{\hat{\bmotion}_M\super{i}}\\[-1mm]
&\cdot\prod_{t=M\shortplus1}^{T}\hspace{-2mm}\q[\textrm{MOT}]{\bposition_t\super{i}}{\hat{\bposition}_t\super{i},\bposition\tm\super{i},\bmotion\tm\super{i}}\\
&\cdot \q[\textrm{MOT}]{\bmotion_t\super{i}}{\hat{\bmotion}_t\super{i}\!\left(\hat{\bposition}\super{i}\ts,\bdescription\super{i},\bsize\super{i}\right),\bposition\tm\super{i},\bmotion\tm\super{i}},
\end{align*}
where $\q[\textrm{AIR}], \q[\textrm{RECT}], \q[\textrm{FIND}],$ and $\q[\textrm{MOT}]$ work as described in \cref{sub:air,sub:find,sub:rect,sub:mot}.
{Variables with hats are intermediate quantities for keeping the model description reasonably short.
	They can be interpreted as point mass distributions without priors.}
\section{Implementation Details}\label{app:implementation}
Detailed algorithmic design of different components of \ac{ours} and the model as a whole is listed in \cref{alg:air_inference,alg:vtssi,alg:air_generation,alg:find,alg:mot,alg:rect,alg:vtssi}. Several points should be taken into account to facilitate the reading of the pseudo code listings:

\begin{itemize}
	\item All random variables mentioned in the listings are Normal, hence parameterized by location (or mean) $\mu$ and scale $\sigma$ (sometimes collectively referred to as ``parameters''). All multivariate Normal distributions are parameterized with diagonal covariance matrices.
	
	\item When a random variable instance results from a computational block, e.g., $\bdescription\super{i} = \operatorname{VAE}_{enc}( \bobs\super{i}_{att} )$, this denotes that the block's output carries the parameters of this random variable. Unless specified otherwise, a computational block produces a concatenated vector of the appropriate size that is then split into required parameter values.
	
	\item Unless specified otherwise, when a random variable is shown as an argument of a computational block (e.g., $\bglimpse\super{i}_{att} = \operatorname{VAE}_{dec}( \bdescription\super{i}) $), a sample taken from that random variable is meant to be fed as an input to the block. 
	The sampling step is not shown in the algorithms to avoid notational clutter.
	
	\item If a sample from the same random variable is used in different parts of an algorithm or different algorithms, in implementation this is a single sample taken once and used multiple times at different parts of a computational flow.
	
	\item Square brackets denote concatenation of multiple vectors inscribed in them.
	
	\item The arguments of the recurrent nets are shown with the running $t$ index that distinguishes between inputs at different time steps to the RNN.
	
	\item The symbol $\odot$ denotes point-wise multiplication.
\end{itemize}

\subsection{Details of AIR Implementation}
\label{app:air}
\label{app:modifications}
The implementation of AIR may be broken down into two major parts: inference model and generative model. 
Here we present the details of our extended AIR implementation, including the modifications described in the main text: namely, position regularization and continuous counting.

\begin{algorithm*}
	\caption{AIR Inference}
	\label{alg:air_inference}
	\DontPrintSemicolon
	\SetKwInOut{Input}{Input}
	\SetKwInOut{Output}{Output}
	\SetSideCommentLeft
	\Input{
		$\bobs$ - single frame, $N$ - maximum number of objects
	}
	
	$\cnt = \operatorname{CNN}_{cnt}\left( \bobs \right)$  \tcp*{object count latent variable}
	$\tilde{n} = N * \operatorname{sigmoid}\left( \cnt \right)$  \tcp*{float number of objects}
	$n = \lceil n \rceil$  \tcp*{int ceiling number of objects}
	
	$\mathbf{f} = \operatorname{CNN}_{pre}\left( \bobs \right)$  \tcp*{frame pre-processing}
	$\left\{\bsize\super{i}, \bposition\super{i}\right\}_{i=1}^{n} = \operatorname{LSTM}_{loc}\left(\{\mathbf{f}\}_{t=1}^{n} \right)$  \tcp*{size and position latent variables}
	
	\For{$i \in [1, \ldots, n]$}{
		$\bobs\super{i}_{att} = \operatorname{ST}(\bobs, \bsize\super{i}, \bposition\super{i})$  \tcp*{inferred object glimpse}
		$\bdescription\super{i} = \operatorname{VAE}_{enc}\left( \bobs\super{i}_{att} \right)$  \tcp*{description latent variable}
	}
	
	\Output{
		$\cnt, \{\bsize\super{i}, \bposition\super{i}, \bdescription\super{i}\}_{i=1}^{n}$
	}
\end{algorithm*}

\paragraph{Inference Model}

The inference model of AIR is shown in \cref{alg:air_inference}. First, the count latent variable $\cnt$ is inferred from the frame $\bobs$ by the counting $\operatorname{CNN}_{cnt}$. The sample from $\cnt$ is squashed by a $\operatorname{sigmoid}$ and multiplied by the maximum number of objects $N$ to arrive at the float number of objects $\tilde{n} \in (0, N)$. $\tilde{n}$ is rounded up to the integer upper bound on the number of objects $n$ that allows limiting the downstream computation (during test time, $\tilde{n}$ is rounded properly to obtain the inferred integer number of objects). Next, the frame $\bobs$ is preprocessed by the $\operatorname{CNN}_{pre}$ and then fed at $n$ time steps to the localization $\operatorname{LSTM}_{loc}$ that outputs the size and position latent variables for every one of the $n$ objects. For every object, a fixed-size glimpse is cropped by the spatial transformer $\operatorname{ST}$ from the frame in accordance with the object's inferred size and position. The inferred glimpse is then encoded by the encoder $\operatorname{VAE}_{enc}$ into a description latent variable of the object. The inference results in a single count variable $\cnt$ and size, position, and description variables $\bsize\super{i}, \bposition\super{i}, \bdescription\super{i}$ for each of the $n$ objects. All latent variables except $\bdescription\super{i}$ have a distinct interpretation.

\begin{algorithm*}
	\caption{AIR Generation}
	\label{alg:air_generation}
	\DontPrintSemicolon
	\SetKwInOut{Input}{Input}
	\SetKwInOut{Output}{Output}
	\SetSideCommentLeft
	\Input{
		$\cnt$ - count latent, $\left\{\bsize\super{i}, \bposition\super{i}, \bdescription\super{i}\right\}_{i=1}^{n}$ - object latents: sizes, positions, and descriptions, \\
		$\sigma_{K}$ - scale of the position regularization bell curve, $\sigma_{L}$ - likelihood scale
	}
	
	$\tilde{n} = N * \operatorname{sigmoid}\left( \cnt \right)$  \tcp*{float number of objects}
	$n = \lceil n \rceil$  \tcp*{int ceiling number of objects}
	$\left\{\text{step}\super{i}\right\}_{i=1}^{n} = \operatorname{split}\left( \tilde{n} \right)$  \tcp*{split $\tilde{n}$ into $n$ $1$-steps (e.g. $\tilde{n}=2.4$ into $[1, 1, 0.4]$)}
	
	$\mu_{L} = 0$  \tcp*{likelihood mean}
	
	$\mathbf{k} = \mathcal{N}(\operatorname{meshgrid}([-1, 1]^2) \mid (0, 0), (\sigma_{K}, \sigma_{K}))$\tcp{discrete regularization kernel, discretization such that $\mathbf{k}$ and $\bglimpse\super{i}_{att}$ are of equal size.}
	
	$\mathbf{k} = \mathbf{k} / \max(\mathbf{k})$ \tcp{normalize kernel to only scale down}
	\For{$i \in [1, \ldots, n]$}{
		$\bglimpse\super{i}_{att} = \operatorname{VAE}_{dec}\left( \bdescription\super{i} \right)$  \tcp*{generated object glimpse}
		$\tilde{\bglimpse}\super{i}_{att} = \bglimpse\super{i}_{att} \odot \mathbf{k}$  \tcp*{position regularization}
		$\hat{\bglimpse}\super{i}_{att} = \tilde{\bglimpse}\super{i}_{att} * \text{step}\super{i}$  \tcp*{continuous counting}
		$\mu\super{i}_{L} = \operatorname{ST^{-1}}\left( \hat{\bglimpse}\super{i}_{att}, \bsize\super{i}, \bposition\super{i} \right)$  \tcp*{partial likelihood mean}
		$\mu_{L} = \mu_{L} + \mu\super{i}_{L}$
	}
	
	$\hat{\bobs} = \mathcal{N}\left( \mu_{L}, \sigma_{L} \right)$  \tcp*{likelihood}
	
	\Output{
		$\hat{\bobs}$
	}
\end{algorithm*}

\paragraph{Generative Model}

The generative model of AIR is shown in \cref{alg:air_generation}. The count variable $\cnt$ is converted into $\tilde{n}$ and $n$ the same way as in the inference model. Next, the float number of objects $\tilde{n}$ is split into a list of step values: consecutive $1$'s totaling to the integer part of $\tilde{n}$ followed by its single fractional remainder (e.g., $\tilde{n}=2.4$ is split into the list $[1, 1, 0.4]$). Next, a generative glimpse of each object is decoded from its description $\bdescription\super{i}$ by the decoder $\operatorname{VAE}_{dec}$. The glimpse then undergoes two consecutive transformations corresponding to position regularization and continuous counting. For position regularization, the glimpse is multiplied by a zero-mean Gaussian bell curve sampled at a uniform grid corresponding to the glimpse pixels in the region $[0, 1]^{2}$. The scale of the bell curve $\sigma_{K}$ is a hyperparameter of the model. The intuition behind the position regularization is that the intensities of the pixels at the center of a glimpse start having more effect on the final generation than the ones closer to the borders of a glimpse. This effect prompts the model to infer the object positions (corresponding to the glimpse centers) closer to the centers of the object "pixel mass", in the attempt to place the majority of object pixels in the high-influence zone of an object glimpse. For continuous counting, each of the glimpses in a row is multiplied by the respective step value resulting from splitting $\tilde{n}$. These step values are used to modulate the effect of each consecutive object on the generated frame: every object except the last one makes it fully into the generation, whereas the effect of the last object is partial as determined by the magnitude of the last fractional step. As the fractional remainder of $\tilde{n}$ is differentiable, guided by the gradient signal through the remainder, the model learns to infer the appropriate number of objects in the frame. Lastly, the resulting glimpse is back-transformed into the original frame dimensions by inverse spatial transformer $\operatorname{ST^{-1}}$ using the object size $\bsize_{i}$ and position $\bposition_{i}$, and pasted onto the cumulative likelihood mean $\mu_{L}$. The final mean $\mu_{L}$ and the fixed scale $\sigma_{L}$ parameterize the output likelihood $\hat{\bobs}$ of the generative model. The scale $\sigma_{L}$ is a model hyperparameter.

\paragraph{Hyperparameters}

$\operatorname{CNN}_{cnt}$ consists of three conv. layers with $16$ $5$x$5$, $4$x$4$, and $3$x$3$ kernels respectively, with ReLU non-linearity applied after convolution. $2$x$2$ max-pooling with strides of $2$ is applied after the first and the second conv. layer. The result is flattened and processed by two dense layers with $256$ and $128$ units and ReLU non-linearity before being linearly transformed to the location and scale of $\cnt$. Before being fed to $\operatorname{CNN}_{cnt}$, a frame is zero-padded with three pixels from each side. $\operatorname{CNN}_{pre}$ consists of two conv. layers with $16$ $3$x$3$ kernels with ReLU non-linearity, each followed by a $2$x$2$ max-pooling layer with stride $2$. The result of $\operatorname{CNN}_{pre}$ is flattened and repetitively fed to $\operatorname{LSTM}_{loc}$ at $n$ steps. $\operatorname{LSTM}_{loc}$ has $256$ units. Dropout with the rate of $0.4$ is applied at training time to the output of $\operatorname{LSTM}_{loc}$, which is then post-processed by four separate dense layers with $64$ units and ReLU non-linearity to arrive at the location and scale of $\bsize\super{i}$ and $\bposition\super{i}$ for every object. The four dense layers are shared between different objects (at $n$ time steps). $\operatorname{VAE}_{enc}$ and $\operatorname{VAE}_{dec}$ are implemented as feed-forward nets with two dense layers with ReLU non-linearity. $\operatorname{VAE}_{enc}$'s layers have $256$ and $128$ units, whereas $\operatorname{VAE}_{dec}$'s layers have $128$ and $256$. The output of $\operatorname{VAE}_{dec}$ is reshaped to the fixed-sized glimpse and taken through $\operatorname{sigmoid}$ (to constrain pixel intensities into $[0, 1]$).

The models are trained with the maximum number of objects $N=2$ (training with $N=3$ did not make a difference when there are no more than $2$ objects in each frame). The glimpse shape is fixed to $25$x$25$ pixels. The size and position variables ($\bsize\super{i}$ and $\bposition\super{i}$) have 2 dimensions (corresponding to X and Y axis). The spatial transformer $\operatorname{ST}$ assumes the size range of $[0, 1]$ ($1$ corresponds to the whole frame) and the position range of $[-1, 1]$ ($-1$ and $1$ correspond to the edges of the frame). To comply with this assumption, the means of $\bsize\super{i}$ and $\bposition\super{i}$ resulting from $\operatorname{LSTM}_{loc}$ are taken through $\operatorname{sigmoid}$ and $\operatorname{tanh}$ respectively. The description variable $\bdescription\super{i}$ is $20$-dimensional. The likelihood scale $\sigma_{L}$ is set to $0.3$. The prior $\p{\bsize\super{i}}$ is a Normal with the location $(0.3, 0.4)$ and the scale $0.1$. The priors $\p{\bposition\super{i}}$ and $\p*	{\bdescription\super{i}}$ are standard normals. The prior $\p{\cnt\super{i}}$ has initial location of $-2.0$ linearly annealed to $-3.0$ between $100$k and $200$k gradient steps, and the scale of $1.0$ (negative locations of $\p{\cnt\super{i}}$ are necessary to mitigate the observed over-counting tendency of AIR). The scale of the position regularization bell curve $\sigma_{K}$ is initially set to $0.5$, but the initial bell curve $K$ is gradually flattened at $1$ during the training. The flattening schedule is $K(t) = (K + p) / (1 + p)$ with the flattening parameter $p$ being linearly annealed from $0.0$ to $100.0$ at the increments of $0.1$ after every $1$k gradient steps. At test time, the position regularization is not applied.

\subsection{Details of VTSSI Implementation}

VTSSI relies on the inference and generative models of AIR described in the previous section as basic building blocks. The components of VTSSI are introduced into the architecture between the inference and generative model of AIR. Below we first describe the details of each of the components and then the whole model formulated in terms of those components.

\begin{algorithm*}
	\caption{FIND}
	\label{alg:find}
	\DontPrintSemicolon
	\SetKwInOut{Input}{Input}
	\SetKwInOut{Output}{Output}
	\SetSideCommentLeft
	\Input{
		$\bobs_{t:T}$ - sequence of frames starting from time $t$, $\bdescription\super{i}$ - object description variable, \\ $\bposition\super{i}_{t-1}$ - object position variable at the previous frame (at time $t-1$)
	}
	
	$\mathbf{k}\super{i} = \operatorname{MLP}_{ker}\left( \bdescription\super{i} \right)$  \tcp*{conv. kernels from the description $\bdescription\super{i}$}
	
	\For{$i \in [t, \ldots, T]$}{
		$\mathbf{f}\super{i}_{t} = \operatorname{CNN}_{find}\left( \bobs_{t}, \mathbf{k}\super{i} \right)$ \tcp*{ features from convolving $\bobs_{t}$ with $\mathbf{k}\super{i}$}
		$\bposition\super{i}_{t} = \operatorname{MLP}_{pos}\left( \left[ \mathbf{f}\super{i}_{t}, \bposition\super{i}_{t-1} \right] \right)$ \tcp*{object position at time $t$}
	}
	
	\Output{
		$\bposition\super{i}_{t:T}$
	}
\end{algorithm*}

\subsubsection{FIND}

FIND is aimed at \emph{tracking} observed objects at future frames. To this end, the latent object description $\bdescription\super{i}$ inferred from the past frame(s), together with the object's previous position $\bposition\super{i}_{t-1}$, are used to discover the object in the current frame $\bobs_{t}$. \Cref{alg:find} depicts an implementation of FIND applied to a sequence of consecutive frames, but it is also straightforward to formulate FIND applied to a single frame (as shown in the \cref{fig:arch-AIR-FIND-RECT,fig:arch-AIR-FIND-MOT}). In contrast to the inference model of AIR that infers object position followed by description, FIND infers new position given description.

\paragraph{Architecture}

The object description $\bdescription\super{i}$ is translated to a bank of convolutional kernels $\mathbf{k}\super{i}$ by $\operatorname{MLP}_{ker}$. The output of $\operatorname{MLP}_{ker}$ is sized and reshaped in accordance with the required number, height, width, and channels in the kernels (which, except channels, are model hyperparameters). As the conv. kernels depend only on the object description that is assumed to be static (and not dynamic), they are computed only once for efficiency and reused with different frames afterwards. The intuition behind $\operatorname{MLP}_{ker}$ is that it \emph{translates} the object description from the latent space to the image space, so that translated description can then be used to find the object at its new position in a new image. Next, the frames of the input sequence $\bobs_{t:T}$ are taken through $\operatorname{CNN}_{find}$ with the first conv. layer parameterized by the conv. kernels $\mathbf{k}\super{i}$, derived from the object description. $\operatorname{CNN}_{find}$ may also have one or more subsequent conv. layers with globally learned weights. Finally, the conv. features $\mathbf{f}\super{i}_{t}$ extracted from the frame $\bobs_{t}$ are concatenated with the object's position at the previous frame $\bposition\super{i}_{t-1}$ and the result is fed through $\operatorname{MLP}_{pos}$ to arrive at the object position variable at the current frame $\bposition\super{i}_{t}$. This process is repeated for every input frame in a row.

\paragraph{Hyperparameters}

$\operatorname{MLP}_{ker}$ consists of two dense layers with 128 and 256 units and ReLU non-linearity. The output of $\operatorname{MLP}_{ker}$ is reshaped into $8$ $10$x$10$ kernels with a single channel. $\operatorname{CNN}_{find}$ consists of a conv. layers parameterized with the kernels derived by $\operatorname{MLP}_{ker}$ following by two globally learned conv. layers with $16$ $5$x$5$ and $32$ $3$x$3$ kernels respectively. $2$x$2$ max-pooling with stride 2 is applied after the first and second conv. layers. The result is flattened and processed by two dense layers with $128$ and $64$ units and ReLU, then linearly transformed into a $50$-dimensional feature vector. $\operatorname{MLP}_{pos}$ consists of two $64$-unit dense layers with tanh non-linearity followed by two separate $32$-unit dense layers, also with tanh non-linearity, and $2$-dimensional linear layers to compute the location and scale of the position variable at the current step $\bposition\super{i}_{t}$. The prior position at the current frame $\p{\bposition\super{i}_{t}}$ is a Normal centered at the (sampled) previous position with the fixed scale of $0.1$. The idea behind this prior is to incorporate an inductive bias of coherent object motion: i.e., the next object position is assumed to be in the neighborhood of the previous one. During training the gradients are not flown through the previous position sample used as a prior mean.

\begin{algorithm*}
	\caption{RECT}
	\label{alg:rect}
	\DontPrintSemicolon
	\SetKwInOut{Input}{Input}
	\SetKwInOut{Output}{Output}
	\SetSideCommentLeft
	\Input{
		$\hat{\cnt}_{1:K}, \left\{\hat{\bsize}\super{i}_{1:K}, \hat{\bdescription}\super{i}_{1:K}\right\}_{i=1}^{N}$ - intermediate object count, size, and description latent variables inferred by AIR from the first $K$ frames $\bobs_{1:K}$ individually
	}
	
	\tcp{feeding concatenated parameters of all intermediate latent variables inferred from the frame $\bobs_{t}$ at the $t$-th time step (variables are shown as arguments instead of parameters to avoid notational clutter)}
	$o_{1:K} = \operatorname{Bi-LSTM}_{rect}\left( \left\{\left[\hat{\cnt}_{t}, \hat{\bsize}\super{1}_{t}, \ldots, \hat{\bsize}\super{N}_{t}, \hat{\bdescription}\super{1}_{t}, \ldots, \hat{\bdescription}\super{N}_{t}\right]\right\}_{t=1}^{K} \right)$ \\
	$w_{1:K} = \operatorname{softmax}\left(o_{1:K} \right)$  \tcp*{rectification weights}
	
	$\cnt = \mathcal{N}\left( \sum_{t=1}^{K} w_{t} * \hat{\cnt}_{t}.\mu, \sum_{t=1}^{K} w_{t}^2 * \hat{\cnt}_{t}.\sigma^2 \right)$  \tcp*{rectified count}
	$n = \lceil N * \operatorname{sigmoid} \left( \cnt \right) \rceil$  \tcp*{rectified ceiling number of objects}
	
	\For{$i \in [1, \ldots, n]$}{
		$\bsize\super{i} = \mathcal{N}\left( \sum_{t=1}^{K} w_{t} * \hat{\bsize}\super{i}_{t}.\mu, \sum_{t=1}^{K} w_{t}^2 * \hat{\bsize}\super{i}_{t}.\sigma^2 \right)$  \tcp*{rectified size}
		$\bdescription\super{i} = \mathcal{N}\left( \sum_{t=1}^{K} w_{t} * \hat{\bdescription}\super{i}_{t}.\mu, \sum_{t=1}^{K} w_{t}^2 * \hat{\bdescription}\super{i}_{t}.\sigma^2 \right)$  \tcp*{rectified description}
	}
	
	\Output{
		$\cnt, \left\{\bsize\super{i}, \bdescription\super{i}\right\}_{i=1}^{n}$
	}
\end{algorithm*}

\subsubsection{RECT}

When objects in a frame are substantially overlapping or partially present, AIR fails to infer an adequate latent representation of the objects. RECT is aimed at \emph{rectifying} potentially incomplete or contaminated latent variables inferred by AIR from multiple frames into a robust object representation. \Cref{alg:rect} shows the details of RECT implementation.

\paragraph{Architecture}

As an input, RECT receives count, size, and description latent variables inferred by AIR from each of the first $K$ frames $\bobs_{1:K}$ individually. The goal is to arrive from those $K$ sets of intermediate variables to a single robust set. RECT solves this tasks by weighted averaging of each variable over the $K$ sets. The $K$ scalar weights used to average every variable are computed by $\operatorname{Bi-LSTM}_{rect}$, to which the concatenated parameters (locations and scales) of all variables in each of the $K$ intermediate sets are fed at $K$ time steps. It is worth mentioning that the input dimensionality of $\operatorname{Bi-LSTM}_{rect}$ at each time step must be fixed to the same number by design. As a consequence, an input at each time step must be concatenated from the same number of latents, which is problematic given the different number of objects that AIR can infer from different frames. To overcome this, either AIR can infer the maximum possible number $N$ of objects from each frame (together with the count variable $\hat{\cnt}$ controlling the effective number of objects), or the parameters of the missing objects variables at different time steps can be replaced by zeros. In our experiments, we adopted the former approach.

The resulting normalized scalar weights $w_{1:K}$ are used first to rectify the intermediate count variables $\hat{\cnt}_{1:K}$ into $\cnt$. Having determined the count, we can proceed with defining the ceiling number of objects $n$ (as in inference and generative models of AIR described above). Finally, we rectify the size and description variables of the $n$ objects by averaging over the respective intermediate variables in the $K$ sets. It is important, that we perform weighted averaging of random variables and not their samples. Resulting single set of rectified latent variables $\cnt, \left\{\bsize\super{i}, \bdescription\super{i}\right\}_{i=1}^{n}$ forms the output of RECT. At test time, one may opt for turning the weights $w_{1:K}$ into a one-hot representation, which amounts to picking a single frame and using exactly AIR-inferred latents from that frame as the rectified ones. However, we have noticed that allowing RECT to combine partial information from different frames leads to more robust rectification (cf. \cref{fig:qualitative-prediction}).

\paragraph{Hyperparameters}

The forward and backward parts of $\operatorname{Bi-LSTM}_{rect}$ both have $128$ hidden units. The forward and backward hidden states at each time step are concatenated and the result is post-processed by two $64$-unit dense layers with ReLU non-linearity (the dense layers are shared among different time steps). After being linearly transformed to scalars, the $K$ outputs are taken through $\operatorname{softmax}$ to arrive at the normalized weights $w_{1:K}$. Weighted average of a set of $K$ intermediate Normal random variables is obtained by weighting the means by $w_{1:K}$ and weighting the variances (squared scales) by $w_{1:K}^{2}$ (an independence assumption is made). AIR's priors are used for the rectified latent variables.

\begin{algorithm*}
	\caption{MOT}
	\label{alg:mot}
	\DontPrintSemicolon
	\SetKwInOut{Input}{Input}
	\SetKwInOut{Output}{Output}
	\SetSideCommentLeft
	\Input{
		$\bsize\super{i}, \bdescription\super{i}$ - the size and description latent variables of an object, $\hat{\bposition}\super{i}_{1:T}$ - the position latent variables of the object at all frames $\bobs_{1:T}$ inferred by AIR and/or FIND, \\
		$M$ - seed motion prefix length, $[w\_min, w\_max]$ - averaging weight interval
	}
	
	$\hat{\bmotion}\super{i}_{M:T} = \operatorname{LSTM}_{mot}\left( \left\{ \left[ \bsize\super{i}, \bdescription\super{i}, \bposition\super{i}_{t} \right] \right\}_{t=1}^{T} \right)$  \tcp*{inferred motion latent variables}
	
	$\bposition\super{i}_{1:M} = \hat{\bposition}\super{i}_{1:M}$  \tcp*{seed position latent variables}
	$\bmotion\super{i}_{M} = \hat{\bmotion}\super{i}_{M}$  \tcp*{seed motion latent variable}
	
	\For{$t \in [M+1, \ldots, T]$}{
		$\tilde{\bposition}\super{i}_{t} = \operatorname{TR}_{pos}\left( \left[ \bposition\super{i}_{t-1}, \bmotion\super{i}_{t-1} \right] \right)$  \tcp*{position prediction (transition)}
		$\tilde{\bmotion}\super{i}_{t} = \operatorname{TR}_{mot}\left( \left[ \bposition\super{i}_{t-1}, \bmotion\super{i}_{t-1} \right] \right)$  \tcp*{motion prediction (transition)}
		
		\tcp{final position and motion variables at time $t$ are obtained by weighted averaging of predicted and inferred variable instances (weight is sampled from a pre-defined interval)}
		$w \sim \operatorname{Uniform}\left( w\_min, w\_max \right)$ \\
		$\bposition\super{i}_{t} = \mathcal{N}\left( w * \tilde{\bposition}\super{i}_{t}.\mu + (1-w) * \hat{\bposition}\super{i}_{t}.\mu, w^{2} * \tilde{\bposition}\super{i}_{t}.\sigma^{2} + (1-w)^{2} * \hat{\bposition}\super{i}_{t}.\sigma^{2} \right)$ \\
		
		$\bmotion\super{i}_{t} = \mathcal{N}\left( w * \tilde{\bmotion}\super{i}_{t}.\mu + (1-w) * \hat{\bmotion}\super{i}_{t}.\mu, w^{2} * \tilde{\bmotion}\super{i}_{t}.\sigma^{2} + (1-w)^{2} * \hat{\bmotion}\super{i}_{t}.\sigma^{2} \right)$		
	}
	
	\Output{
		$\bposition\super{i}_{1:T}$, 
		$\bmotion\super{i}_{M:T}$,
		$\tilde{\bposition}\super{i}_{M+1:T}$,
		$\tilde{\bmotion}\super{i}_{M+1:T}$
	}
\end{algorithm*}

\subsubsection{MOT}

The components described so far are targeted at inferring the latent representation from the available observations. AIR is capable of understanding a scene, FIND can reliably track the objects seen before, RECT can disentangle object representations. But none of those components is able to predict the future given the observed past. MOT is introduced to fill in this gap, as it includes a state-space model of object motion.

\paragraph{Architecture} To model the motion, MOT introduces a new latent variable -- $\bmotion\super{i}_{t}$ -- describing the motion of $i$-th object at the $t$-th frame. Albeit not interpretable, this motion description can in principle carry information about object velocity or other higher-order characteristics of the motion. Architecturally, MOT consists of two components: one aimed at inferring $\bmotion\super{i}_{t}$ from a sequence of past object positions and the other being able to predict the future object position and motion variables given those at the current time step.

As an input, MOT receives a sequence of object positions $\hat{\bposition}\super{i}_{1:T}$ inferred from all frames of the sequence (e.g., by FIND), alongside the size $\bsize\super{i}$ and the description $\bdescription\super{i}$ of the object. As the first step, MOT infers the motion variables $\hat{\bmotion}\super{i}_{M:T}$ at all time steps starting from $M$-th (the positions at the first $M$ steps $\bposition\super{i}_{1:M}$ are used to gain initial awareness of the motion pattern, hence the motion variables are inferred starting from the $M$-th step). This is achieved by feeding the positions at the time steps from $1$ to $T$, each concatenated with the object size and description, at $T$ time steps to $\operatorname{LSTM}_{mot}$. The result is a sequence of inferred motion variables $\hat{\bmotion}\super{i}_{M:T}$ (due to the reasons described above, the $\operatorname{LSTM}_{mot}$ outputs at the steps before $M$ are ignored).

The inferred position variables $\hat{\bposition}\super{i}_{1:M}$ and motion variable $\hat{\bmotion}\super{i}_{M}$ are treated as the final position and motion variables at those steps $\bposition\super{i}_{1:M}$ and $\bmotion\super{i}_{M}$ respectively. The final position and motion variables at the steps from $M+1$ to $T$ are obtained through the remaining part of MOT: prediction-averaging loop. At every iteration of this loop, starting from the time step $M+1$, the concatenated final position and motion variables at the previous step $\bposition\super{i}_{t-1}$ and $\bmotion\super{i}_{t-1}$ are taken through the position transition network $\operatorname{TR}_{pos}$ and motion transition network $\operatorname{TR}_{mot}$ to arrive at the position prediction $\tilde{\bposition}\super{i}_{t}$ and the motion prediction $\tilde{\bmotion}\super{i}_{t}$ variables at the current time step respectively. Finally, each prediction variable ($\tilde{\bposition}\super{i}_{t}$ and $\tilde{\bmotion}\super{i}_{t}$) is weighted-averaged with the corresponding inferred variable ($\hat{\bposition}\super{i}_{t}$ and $\hat{\bmotion}\super{i}_{t}$) to obtain the final variable at time step $t$ ($\bposition\super{i}_{t}$ and $\bmotion\super{i}_{t}$). Averaging weight $w \in [0, 1]$ is sampled from a uniform distribution with the predefined minimum and maximum bounds. Those bounds can be changed in the course of training to regularize and/or control the relative effect of prediction and inference on the final position and motion variables.

Position and motion transition networks -- $\operatorname{TR}_{pos}$ and $\operatorname{TR}_{mot}$ -- jointly comprise the state-space model of MOT. By applying the transition networks repetitively, one can perform fully generative sampling of future object positions, hence predict future object motion conditioned on the past.

\paragraph{Hyperparameters}

$\operatorname{LSTM}_{mot}$ has $64$ hidden units. The hidden state at each time step is post-processed by two separate $32$-unit dense layers with tanh non-linearity, followed by linear transformations to compute the location and scale of the inferred motion variables $\hat{\bmotion}\super{i}_{M:T}$. Each of the two transition networks consists of two $64$-unit dense layers with tanh non-linearity, followed by two separate $32$-unit dense layers with tanh non-linearity, followed by linear transformations to compute the location and scale of the predicted variable.

The motion variable $\bmotion\super{i}_{t}$ is a $10$-dimensional Normal random variable. The prior $\p{\bmotion\super{i}_{M}}$ is a standard Normal. At the steps from $M+1$ to $T$, the position prediction $\tilde{\bposition}\super{i}_{t}$ and motion prediction $\tilde{\bmotion}\super{i}_{t}$ variables are used as priors for the final position $\bposition\super{i}_{t}$ and final motion $\bmotion\super{i}_{t}$ variables respectively. During training, the averaging weight $w$ is sampled from the $\operatorname{Uniform}\left[ 0.01, 0.99 \right]$; during test time the weight is fixed to $0.5$.

\begin{algorithm*}
	\caption{VTSSI}
	\label{alg:vtssi}
	\DontPrintSemicolon
	\SetKwInOut{Input}{Input}
	\SetKwInOut{Output}{Output}
	\SetSideCommentLeft
	\Input{
		$\bobs_{1:T}$ - sequence of frames, $N$ - maximum number of objects, \\
		$K$ - rectification prefix length, $M$ - seed motion prefix length
	}
	
	$\hat{\cnt}_{1:K}, \left\{\hat{\bsize}\super{i}_{1:K}, \hat{\bdescription}\super{i}_{1:K}\right\}_{i=1}^{N} = \left\{\operatorname{AIR}_{inf}\left( \bobs_{t}, N \right) \right\}_{t=1}^{K}$  \tcp*{intermediate variables}
	
	$\cnt, \left\{ \bsize\super{i}, \bdescription\super{i} \right\}_{t=1}^{n} = \operatorname{RECT}\left( \hat{\cnt}_{1:K}, \left\{\hat{\bsize}\super{i}_{1:K}, \hat{\bdescription}\super{i}_{1:K}\right\}_{i=1}^{N} \right)$  \tcp*{rectified variables}
	
	\For{$i \in [1, \ldots, n]$}{
		$\hat{\bposition}\super{i}_{1:T} = \operatorname{FIND}\left( \bobs_{1:T}, \bdescription\super{i}, \hat{\bposition}\super{i}_{0}=0 \right)$  \tcp*{position inference}
		
		$\bposition\super{i}_{1:T}, \bmotion\super{i}_{M:T}, \tilde{\bposition}\super{i}_{M+1:T}, \tilde{\bmotion}\super{i}_{M+1:T} = \operatorname{MOT}\left( \bsize\super{i}, \bdescription\super{i}, \hat{\bposition}\super{i}_{1:T}, M \right)$  \tcp*{state-space model}
	}
	
	$\hat{\bobs}_{1:T} = \left\{ \operatorname{AIR}_{gen}\left( \cnt, \left\{\bsize\super{i}, \bdescription\super{i}, \bposition\super{i}_{t}\right\}_{i=1}^{n} \right) \right\}_{t=1}^{T}$  \tcp*{likelihood}
	
	\Output{
		$\hat{\bobs}_{1:T}, \cnt, \left\{\bsize\super{i}, \bdescription\super{i}, \bposition\super{i}_{1:T}, \bmotion\super{i}_{M:T}, \tilde{\bposition}\super{i}_{M+1:T}, \tilde{\bmotion}\super{i}_{M+1:T}\right\}_{i=1}^{n}$
	}
\end{algorithm*}

\subsubsection{VTSSI}

VTSSI relies on the components described above for accomplishing higher-level task. 
First, AIR inference model is run on the first $K$ frames $\bobs_{1:K}$ separately to infer intermediate object counts, sizes, and descriptions. 
Next, the intermediate variables are rectified into the final count, size, and description variables by RECT. 
Next, for every object, FIND is used to infer the object positions at every frame in the sequence $\bobs_{1:T}$ (zero is fed instead of the initial position to FIND, as the initial position is unavailable for a rectified object). 
Next, the object motion is modeled with MOT that produces the final position and motion variables. 
Lastly, the rectified object count, sizes, and descriptions, alongside the final positions, are fed to AIR generative model to obtain the likelihood at every time step $\hat{\bobs}_{1:T}$.
When a tracking task does not pose the challenges of disentangling and/or prediction, RECT and/or MOT can be trivially excluded from the architecture to reduce overall model complexity.
In the experiments reported in \cref{tab:results} and \cref{fig:qualitative-comparison-ours}, the models were trained with $N=2$, $K=5$ and $M=5$ on sequences of length $T=20$.
Component-specific hyperparameters are described in the corresponding component subsections above.
The differences in hyperparameters used for training the models for baseline comparisons are described in a separate section below.

\section{Training}

The model is trained by maximizing ELBO using Adam optimizer with $\beta_{1}=0.5$ and mini-batches of $64$ sequences.
The training lasts for $1$k epochs, which approximately corresponds to $780$k gradient steps.
On our hardware setup, this amounts to a wall-clock time of roughly 83 hours.
The learning rate is initialized at $1e-4$ and smoothly annealed down to $1e-5$ starting after $200$k gradient steps at the rate of $0.9$ per $20$k gradient steps.
Gradients are clipped (by global norm) at $5.0$ for higher training stability.

\label{app:curriculum}
\subsection{Curriculum}

Curriculum learning is used to progressively increase the complexity of task as the model trains.
For all models except VTSSI, curriculum starts with the sequences of length $1$ (effectively, training AIR on the first frames), with the length being incremented by $1$ every $20$k gradient steps.
VTSSI is trained with the curriculum starting at $6$ and with the increments of $1$ after every $30$k gradient steps.

While training VTSSI, we found it beneficial for overall training stability to train AIR on the first $K$ frames jointly with the full model (i.e., adding AIR ELBO to the VTSSI ELBO in the loss function) during the first several steps of the curriculum.
In our experiments, AIR is trained in parallel with VTSSI during the first $3$ steps.
Starting from the $4$-th curriculum step, the AIR ELBO term is dropped from the loss function.

\subsection{Baselines}

There were minor differences in the hyperparameter configuration of the models trained for comparison with the baselines (with the results reported in \cref{fig:quantitative-prediction}).
We list those differences here.

\paragraph{DDPAE}

The glimpse shape was set to $[32, 32]$ instead of $[25, 25]$. 
The size of the computed conv. kernels of FIND were set to $[12, 12]$ instead of $[10, 10]$. 
The averaging weight of MOT was fixed at $0.5$ during training (no random sampling). 
Learning rate was annealed down to $5e-5$ instead of $1e-5$.
Curriculum started at the sequences of length $2$ instead of $6$. $K = M = 10$ instead of $5$.

\paragraph{SQAIR}

RECT component was not used, as the first frame was clean.
The mean of the position variables inferred by FIND and predicted by MOT were scaled with by factor of $1.5$ (after taking through tanh) to allow placing objects partially out of frame.
The averaging weight of MOT was sampled from $\operatorname{Uniform}[0.1, 0.9]$ instead of $\operatorname{Uniform}[0.01, 0.99]$ at training time. $K=3$ instead of $5$. $T=10$ instead of $20$.

\section{Experiments}\label{app:experiments}

\subsection{Data set details}

Our datasets consist of $50,000$ training, $10,000$ validation, and $10,000$ test sequences with variable number of MNIST digits moving within $50$x$50$ frames.
The length of the sequences is $20$. 
The number of digits in each sequence is sampled uniformly at random from $\{0, 1, 2\}$, but is fixed for each sequence. 
MNIST digits for each sequence are sampled uniformly at random from the original MNIST dataset. 
The MNIST digits in our test set are sampled only from the MNIST test set, whereas the ones in our training and validation sets are sampled only from the MNIST training set.

Four versions of our dataset are determined by combination of two factors: 
\begin{itemize}
	\item whether digit motion is linear or elliptic
	\item whether two digits in the first frame are allowed to overlap
\end{itemize}

The digits are placed at random position in the initial frame with the conditions of residing within the frame. 
In non-overlapping first frame dataset two digits are not allowed to overlap in the first frame: i.e., they may not share non-zero intensity pixels (but may still overlap in further frames).

In the dataset with linear motion, random velocity vector is sampled for each digit and kept constant during motion, except flipping the components of the velocity at the edges of the frame: when at least one pixel of the digit goes out of frame after a motion step, the digit bounces off the edge.

In the dataset with elliptic motion, random elliptic trajectory is sampled for each digit such that a digit stays within the frame while moving along it. 
Angular velocity of each individual object is also sampled randomly and kept constant throughout the sequence.

As the velocity magnitudes are sampled from uniform distributions, while objects are moving, their positions take real values. 
Instead of rounding the position to the nearest integer pixel and pasting the same constellation of pixels as in the original digit at a new discrete position, we maintain the real position values and through bilinear interpolation smoothen the digit motion. 
We believe that this makes our datasets closer to real video sequences, where object motion is typically smooth.

\paragraph{DDPAE}

DDPAE and VTSSI models with the prediction performance reported in \cref{fig:quantitative-ddpae} were trained on the data generated by the script from the official DDPAE repository\footnote{\url{https://github.com/jthsieh/DDPAE-video-prediction}}. 
The test set was also generated by the DDPAE script, because the original Moving MNIST dataset lacks ground truth position annotation. 
It is worth mentioning that VTSSI was trained on $50,000$ $20$-frame sequences, whereas DDPAE was trained on streaming data (with every batch being randomly generated). 
The performance of both models reported in \cref{fig:quantitative-ddpae} is evaluated on the test set.

\paragraph{SQAIR}

SQAIR and VTSSI models with the prediction performance reported in \cref{fig:sqair-prediction-sqair,fig:sqair-prediction-lin} were trained on three different datasets corresponding to the two figures. 
SQAIR data corresponding to \cref{fig:sqair-prediction-sqair} was generated by the data generation script from the official SQAIR repository\footnote{\url{https://github.com/akosiorek/sqair}}, without noise and acceleration in digit motion. 
Our linear data corresponding to \cref{fig:sqair-prediction-lin} is comprised of 10-frame sequences structurally similar to our non-overlapping linear dataset, with the exception of all frame edges being virtually shifted 3 pixels away from the center. 
This is to allow the digits going deeper out of frame before bouncing (for higher similarity with SQAIR's data).
Model performance reported in \cref{fig:quantitative-prediction} is evaluated on hold-out test sets.

\label{appsub:data}

\subsection{Evaluation details}

The accuracies reported in \cref{tab:results} are computed by dividing the number of sequences, where the number of objects is correctly inferred by the total number of sequences in the test set. 
AIR's accuracy is computed per-frame, as it may infer different numbers of objects from different frames of a single sequence (e.g., when the objects are highly overlapping).

The position error reported in \cref{tab:results} and \cref{fig:quantitative-prediction,fig:quantitative-ddpae} is computed as a distance in pixels between the ground truth object position (part of the dataset meta-data) and the positions inferred or predicted by the model. 
Ground truth object positions in all datasets correspond to the geometric centers of the tight bounding boxes around the object.
The positions inferred or predicted by the models are translated into pixel coordinates before being compared with the ground truth positions.
The position error is computed per inferred object and not per sequence: i.e., if there are two objects in one sequence, those are treated as two different subjects of comparison.
When there are multiple possible matchings between ground truth and inferred objects, we pick the matching that minimizes the summed distance error on a prefix of a sequence.
Observation horizons of the models are used as the length of matching-determining prefixes (e.g., 10 in VTSSI vs. DDPAE and 3 in VTSSI vs. SQAIR evaluation).

At test time, DDPAE and VTSSI replace random variables in the computational graph by their modes.
This proves to yield more accurate one-shot long-term predictions of object motion.
As the SQAIR code from the official repository samples generative trajectories randomly, this would give a comparative disadvantage to SQAIR.
For this reason, during evaluation we have modified SQAIR code to replace all random variables by their modes, the same way as DDPAE and VTSSI do.
This modification substantially improved the prediction performance metrics of SQAIR.
We also modified the configuration of the trained SQAIR models to avoid dropping the objects from the sequence, even when they disappear behind an edge of a frame. After this change SQAIR always preserved the objects inferred from the first frame throughout the sequence.

\section{Further results}\label{app:results}

\label{appsub:variants-of-ours}
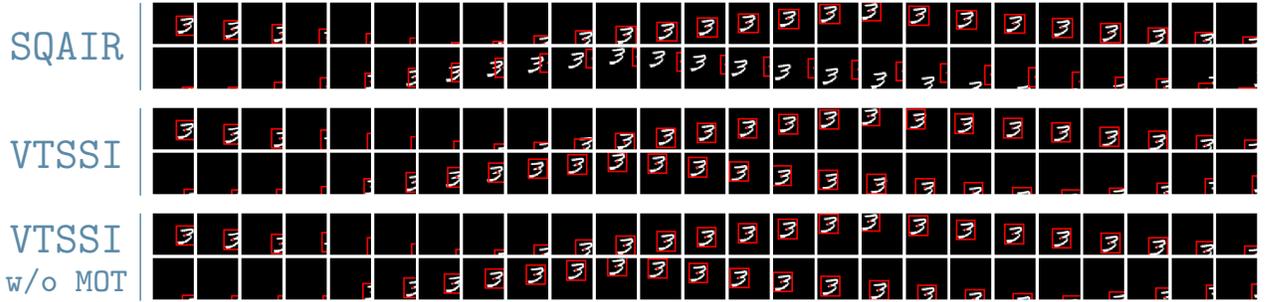
\begin{figure*}[h!]
	\centering
	\begin{subfigure}{\textwidth}
		\resizebox{\textwidth}{!}{
			\begin{tikzpicture}[scale=0.2]
			\path (0, 4) grid (60, 18);
			
			\path [draw, dblue] (3.5, 15.8) node {\large \texttt{SQAIR}};
			\draw [dblue] (7, 18) -- (7, 13.85);
			\node[inner sep=0pt] at (33.8, 15.95) {\includegraphics[width=10.5cm]{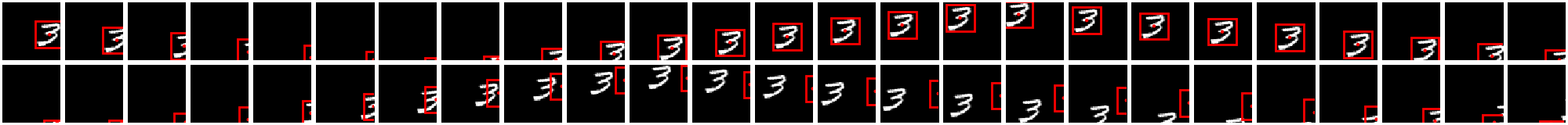}};
			
			\path [draw, dblue] (3.5, 10.8) node {\large \texttt{VTSSI}};
			\draw [dblue] (7, 13) -- (7, 8.85);
			\node[inner sep=0pt] at (33.8, 10.95) {\includegraphics[width=10.5cm]{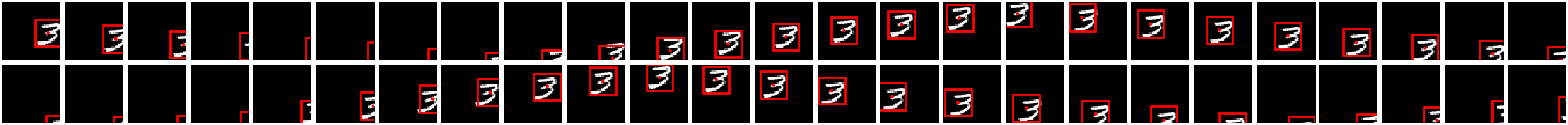}};
			
			\path [draw, dblue] (3.5, 6.8) node {\large \texttt{VTSSI}};
			\path [draw, dblue] (3.5, 4.8) node {\small \texttt{w/o MOT}};
			\draw [dblue] (7, 8) -- (7, 3.85);
			\node[inner sep=0pt] at (33.8, 5.95) {\includegraphics[width=10.5cm]{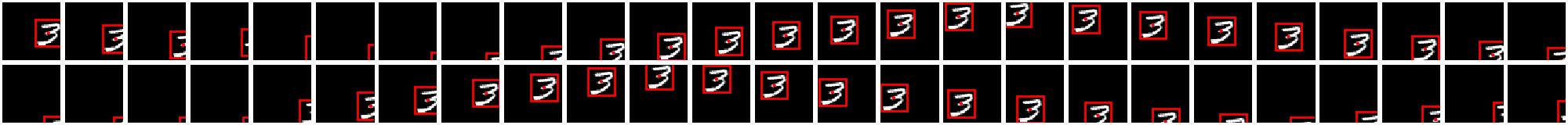}};
			\end{tikzpicture}
		}
	\end{subfigure}
	\caption{
		Tracking performance of SQAIR vs. VTSSI on the same sequence (from the SQAIR dataset).
	}
	\label{fig:sqair-tracking}
\end{figure*}

\begin{figure*}[h!]
	\centering
	\begin{subfigure}{\textwidth}
		\resizebox{\textwidth}{!}{
			\begin{tikzpicture}[scale=0.2]
			\path (0, 9) grid (60, 18);
			
			\path [draw, dblue] (3.5, 15.8) node {\large \texttt{SQAIR}};
			\draw [dblue] (7, 18) -- (7, 13.85);
			\node[inner sep=0pt] at (33.8, 15.95) {\includegraphics[width=10.5cm]{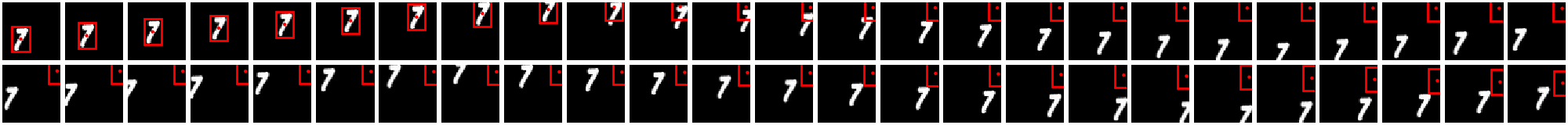}};
			
			\path [draw, dblue] (3.5, 10.8) node {\large \texttt{VTSSI}};
			\draw [dblue] (7, 13) -- (7, 8.85);
			\node[inner sep=0pt] at (33.8, 10.95) {\includegraphics[width=10.5cm]{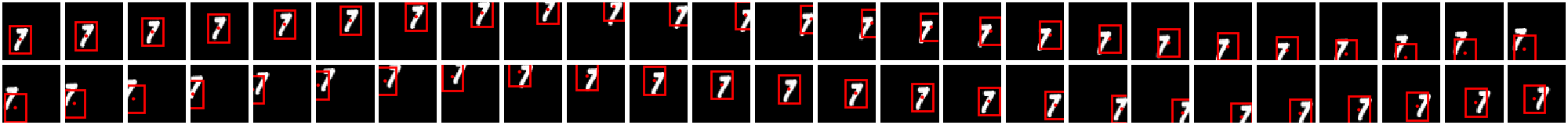}};
			\end{tikzpicture}
		}
	\end{subfigure}
	\caption{
		Prediction performance of SQAIR vs. VTSSI on the same sequence (from our dataset).
	}
	\label{fig:sqair-prediction}
\end{figure*}
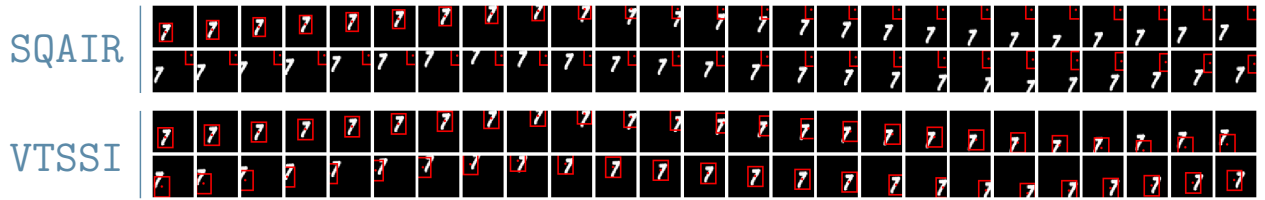

\Cref{fig:sqair-tracking,fig:sqair-prediction} show an example of prediction and tracking of a long sequence, highlighting the findings discussed in \cref{sub:exp-inference,sub:exp-prediction}.

\begin{figure*}[h!]
	\centering
	\begin{subfigure}{\textwidth}
		\resizebox{\textwidth}{!}{
			\begin{tikzpicture}[scale=0.2]
			\path (0, -26) grid (60, 20);
			
			\foreach \x in {1,...,20}
			{
				\path [draw, dblue] (6.3+2.62*\x, 19) node {\tiny \texttt{\x}};
			}
			
			\path [draw, dblue] (3.5, 15) node {\Large \texttt{AIR}};
			\draw [dblue] (7, 18) -- (7, 12.5);
			
			\node[inner sep=0pt] at (33.8, 16.8) {\includegraphics[width=10.5cm]{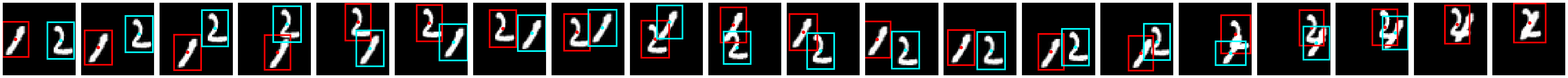}};
			\node[inner sep=0pt] at (33.8, 13.8) {\includegraphics[width=10.5cm]{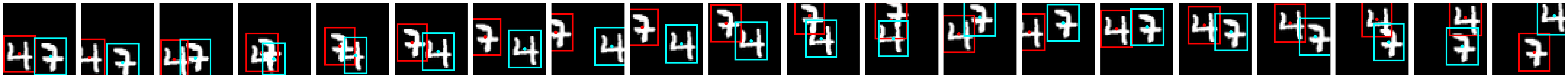}};

			\path [draw, dblue] (3.5, 8) node {\large \texttt{FIND}};
			\draw [dblue] (7, 11) -- (7, 5.5);
			
			\node[inner sep=0pt] at (33.8, 9.8) {\includegraphics[width=10.5cm]{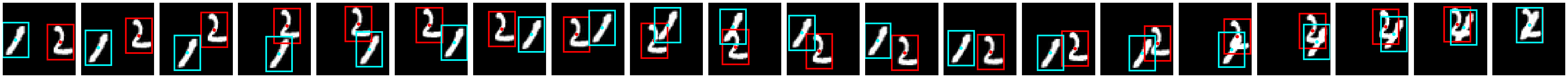}};
			\node[inner sep=0pt] at (33.8, 6.8) {\includegraphics[width=10.5cm]{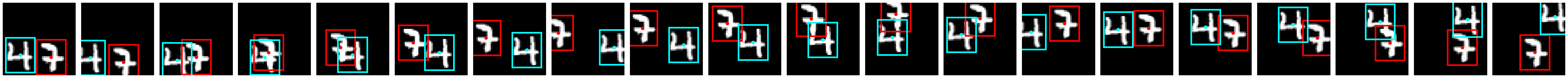}};

			\path [draw, dblue] (3.5, 1) node[text width=1cm, align=center] {\large \texttt{RECT FIND}};
			\draw [dblue] (7, 4) -- (7, -1.5);
			
			\node[inner sep=0pt] at (33.8, 2.8) {\includegraphics[width=10.5cm]{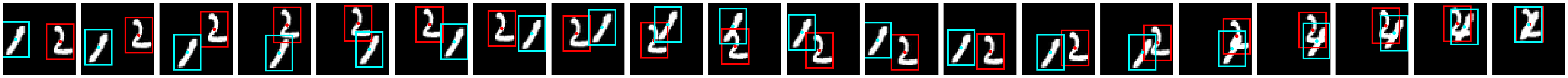}};
			\node[inner sep=0pt] at (33.8, -0.2) {\includegraphics[width=10.5cm]{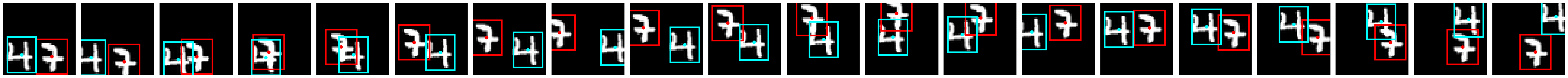}};

			\path [draw, dblue] (3.5, -8) node[text width=1cm, align=center] {\large \texttt{FIND MOT}};
			\draw [dblue] (7, -3) -- (7, -13.7);
			
			\node[inner sep=0pt] at (33.8, -5.5) {\includegraphics[width=10.5cm]{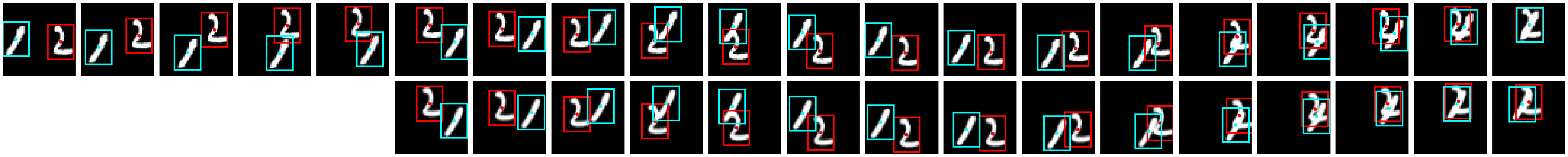}};
			\path [draw, dblue] (12.5, -7) node {\texttt{prediction}};
			\draw [->, >=latex, dblue] (17.8, -6.9) -- (19.8, -6.9);
			\node[inner sep=0pt] at (33.8, -11.1) {\includegraphics[width=10.5cm]{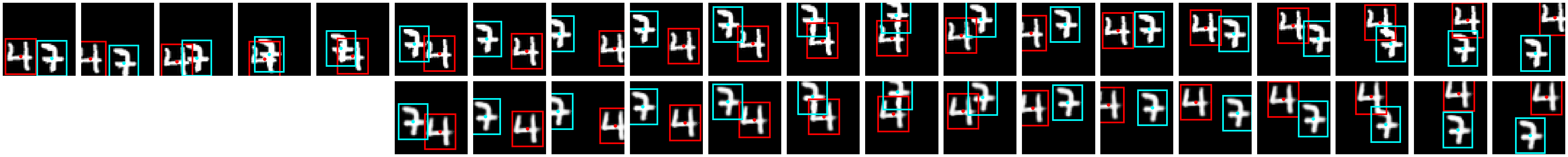}};
			\path [draw, dblue] (12.5, -12.6) node {\texttt{prediction}};
			\draw [->, >=latex, dblue] (17.8, -12.5) -- (19.8, -12.5);

			\path [draw, dblue] (3.5, -20) node {\large \texttt{VTSSI}};
			\draw [dblue] (7, -16) -- (7, -25.7);
			
			\node[inner sep=0pt] at (33.8, -17.5) {\includegraphics[width=10.5cm]{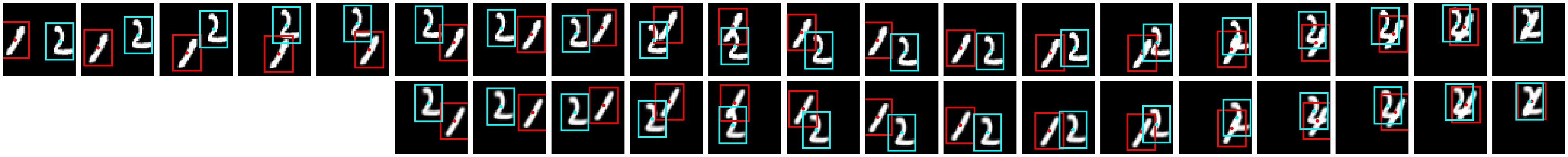}};
			\path [draw, dblue] (12.5, -19) node {\texttt{prediction}};
			\draw [->, >=latex, dblue] (17.8, -18.9) -- (19.8, -18.9);
			\node[inner sep=0pt] at (33.8, -23.1) {\includegraphics[width=10.5cm]{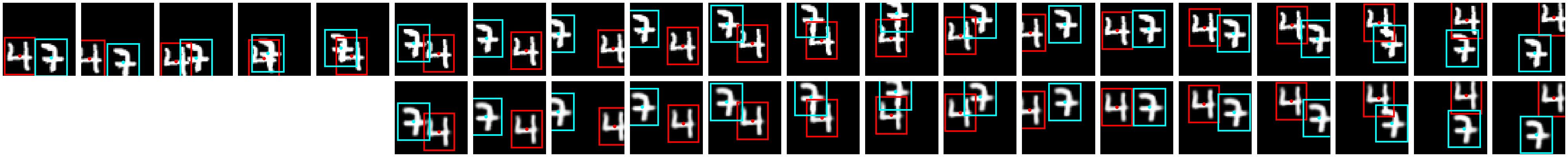}};
			\path [draw, dblue] (12.5, -24.6) node {\texttt{prediction}};
			\draw [->, >=latex, dblue] (17.8, -24.5) -- (19.8, -24.5);
			
			\end{tikzpicture}
		}
	\end{subfigure}
	\caption{
		Sequences with non-overlapping initial frames evaluated an all flavors of \ac{ours}.
	}
	\label{fig:disentangled-sequences}
\end{figure*}

\begin{figure*}[h!]
	\centering
	\begin{subfigure}{\textwidth}
		\resizebox{\textwidth}{!}{
			\begin{tikzpicture}[scale=0.2]
			\path (0, -26) grid (60, 20);
			
			\foreach \x in {1,...,20}
			{
				\path [draw, dblue] (6.3+2.62*\x, 19) node {\tiny \texttt{\x}};
			}
			
			\path [draw, dblue] (3.5, 15) node {\Large \texttt{AIR}};
			\draw [dblue] (7, 18) -- (7, 12.5);
			
			\node[inner sep=0pt] at (33.8, 16.8) {\includegraphics[width=10.5cm]{gfx/sequences/ent/air/0.png}};
			\node[inner sep=0pt] at (33.8, 13.8) {\includegraphics[width=10.5cm]{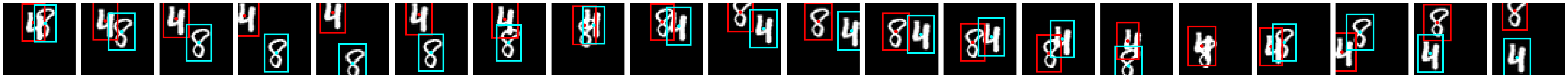}};

			\path [draw, dblue] (3.5, 8) node {\large \texttt{FIND}};
			\draw [dblue] (7, 11) -- (7, 5.5);
			
			\node[inner sep=0pt] at (33.8, 9.8) {\includegraphics[width=10.5cm]{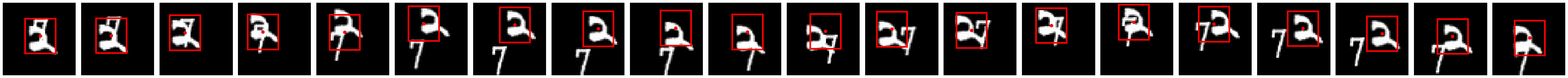}};
			\node[inner sep=0pt] at (33.8, 6.8) {\includegraphics[width=10.5cm]{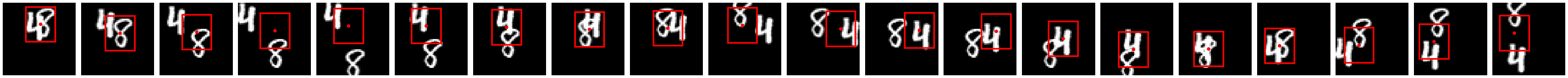}};

			\path [draw, dblue] (3.5, 1) node[text width=1cm, align=center] {\large \texttt{RECT FIND}};
			\draw [dblue] (7, 4) -- (7, -1.5);
			
			\node[inner sep=0pt] at (33.8, 2.8) {\includegraphics[width=10.5cm]{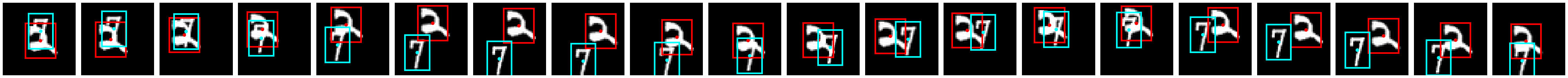}};
			\node[inner sep=0pt] at (33.8, -0.2) {\includegraphics[width=10.5cm]{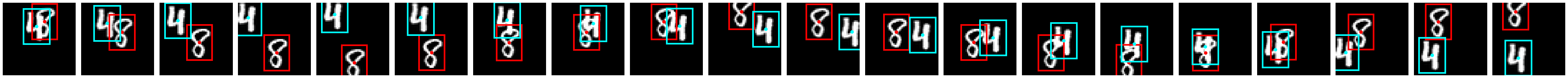}};

			\path [draw, dblue] (3.5, -8) node[text width=1cm, align=center] {\large \texttt{FIND MOT}};
			\draw [dblue] (7, -3) -- (7, -13.7);
			
			\node[inner sep=0pt] at (33.8, -5.5) {\includegraphics[width=10.5cm]{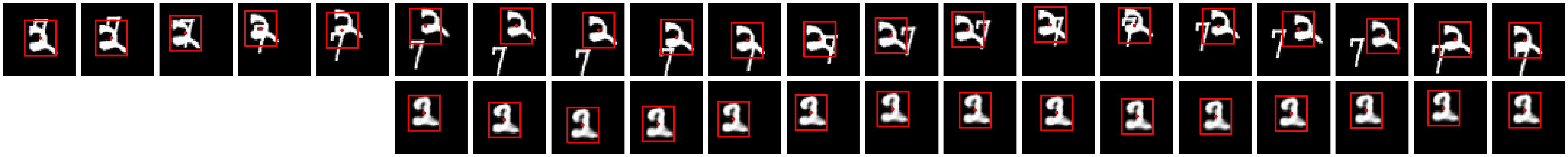}};
			\path [draw, dblue] (12.5, -7) node {\texttt{prediction}};
			\draw [->, >=latex, dblue] (17.8, -6.9) -- (19.8, -6.9);
			\node[inner sep=0pt] at (33.8, -11.1) {\includegraphics[width=10.5cm]{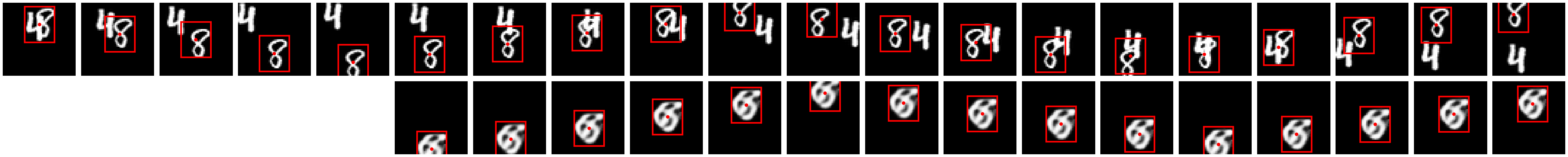}};
			\path [draw, dblue] (12.5, -12.6) node {\texttt{prediction}};
			\draw [->, >=latex, dblue] (17.8, -12.5) -- (19.8, -12.5);

			\path [draw, dblue] (3.5, -20) node {\large \texttt{VTSSI}};
			\draw [dblue] (7, -16) -- (7, -25.7);
			
			\node[inner sep=0pt] at (33.8, -17.5) {\includegraphics[width=10.5cm]{gfx/sequences/ent/air_rect_find_mot/0.png}};
			\path [draw, dblue] (12.5, -19) node {\texttt{prediction}};
			\draw [->, >=latex, dblue] (17.8, -18.9) -- (19.8, -18.9);
			\node[inner sep=0pt] at (33.8, -23.1) {\includegraphics[width=10.5cm]{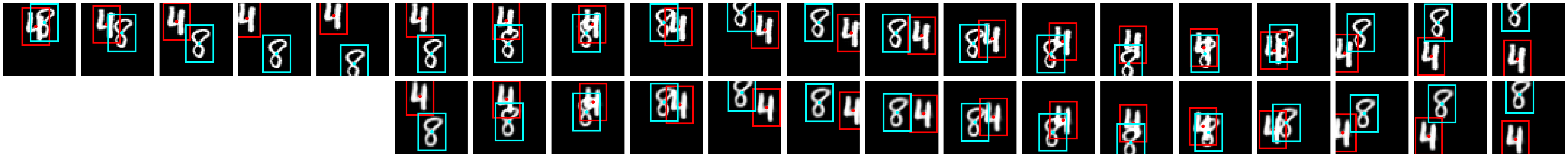}};
			\path [draw, dblue] (12.5, -24.6) node {\texttt{prediction}};
			\draw [->, >=latex, dblue] (17.8, -24.5) -- (19.8, -24.5);
			
			\end{tikzpicture}
		}
	\end{subfigure}
	\caption{
		Sequences with overlapping initial frames evaluated an all flavors of \ac{ours}.
	}
	\label{fig:entangled-sequences}
\end{figure*}

\Cref{fig:disentangled-sequences,fig:entangled-sequences} show further sequences with flavors of \ac{ours}, comparable to \cref{fig:qualitative-comparison-ours}.

\Cref{fig:vtssi-inference} shows inference of \ac{ours} on random sequences.
The top 12 are elliptic, the bottom 12 are with linear motion.

\Cref{fig:vtssi-prediction} shows seeded generative prediction on the same sequences as \cref{fig:vtssi-inference}.
The inference seed horizon is $K=M=5$.
The depicted frames are ground truth, the bounding boxes are superimposed from the predictions.

\Cref{fig:vtssi-prediction-gen-frames} shows the same predictions, but with generated frames instead of ground truth.

\begin{figure*}[h!]
	\centering
	\begin{subfigure}{\textwidth}
		\centering
		\resizebox{!}{\textheight}{
			\begin{tikzpicture}[scale=0.2]
				\foreach \x in {0,...,23}
				{
					\node[inner sep=0pt] at (30, 18.5-4*\x) {\includegraphics[width=12cm]{gfx/sequences/vtssi/inf_real/\x}};
				}
			\end{tikzpicture}
		}
	\end{subfigure}
	\caption{
		Inference of \ac{ours} superimposed on ground truth frames of random sequences.
	}
	\label{fig:vtssi-inference}
\end{figure*}

\begin{figure*}[h!]
	\centering
	\begin{subfigure}{\textwidth}
		\centering
\resizebox{!}{\textheight}{
			\begin{tikzpicture}[scale=0.2]
				\foreach \x in {0,...,23}
				{
					\node[inner sep=0pt] at (30, 18.5-4*\x) {\includegraphics[width=12cm]{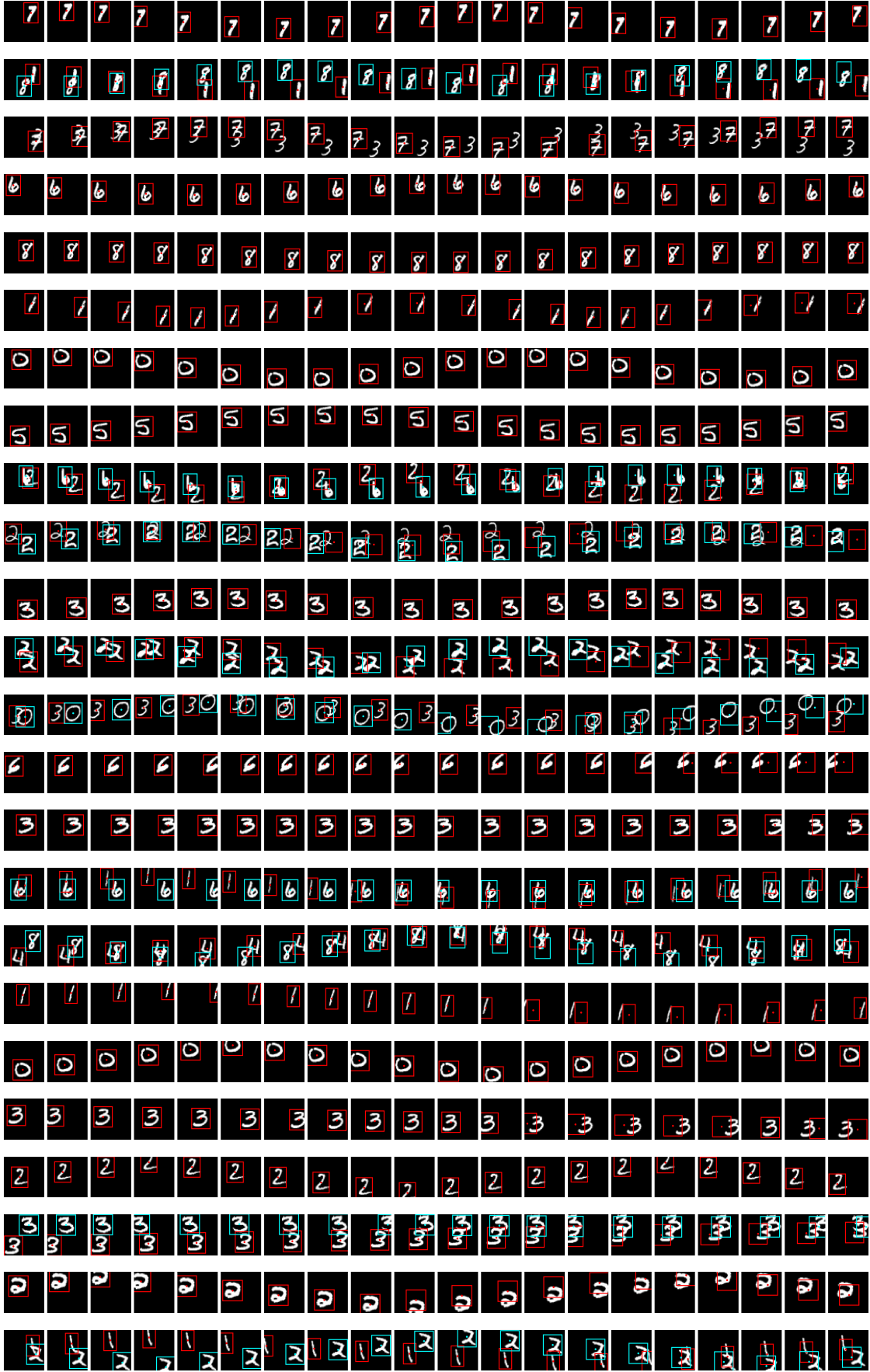}};
				}
			\end{tikzpicture}
		}
	\end{subfigure}
	\caption{
		Generative predictions of \ac{ours} superimposed on ground truth frames of the same random sequences as in \cref{fig:vtssi-inference}.
		Generation is seeded with $K=M=5$ frames of the ground truth.
	}
	\label{fig:vtssi-prediction}
\end{figure*}

\begin{figure*}
	\centering
	\begin{subfigure}{\textwidth}
		\centering
\resizebox{!}{\textheight}{
			\begin{tikzpicture}[scale=0.2]
			\foreach \x in {0,...,23}
			{
				\node[inner sep=0pt] at (30, 18.5-4*\x) {\includegraphics[width=12cm]{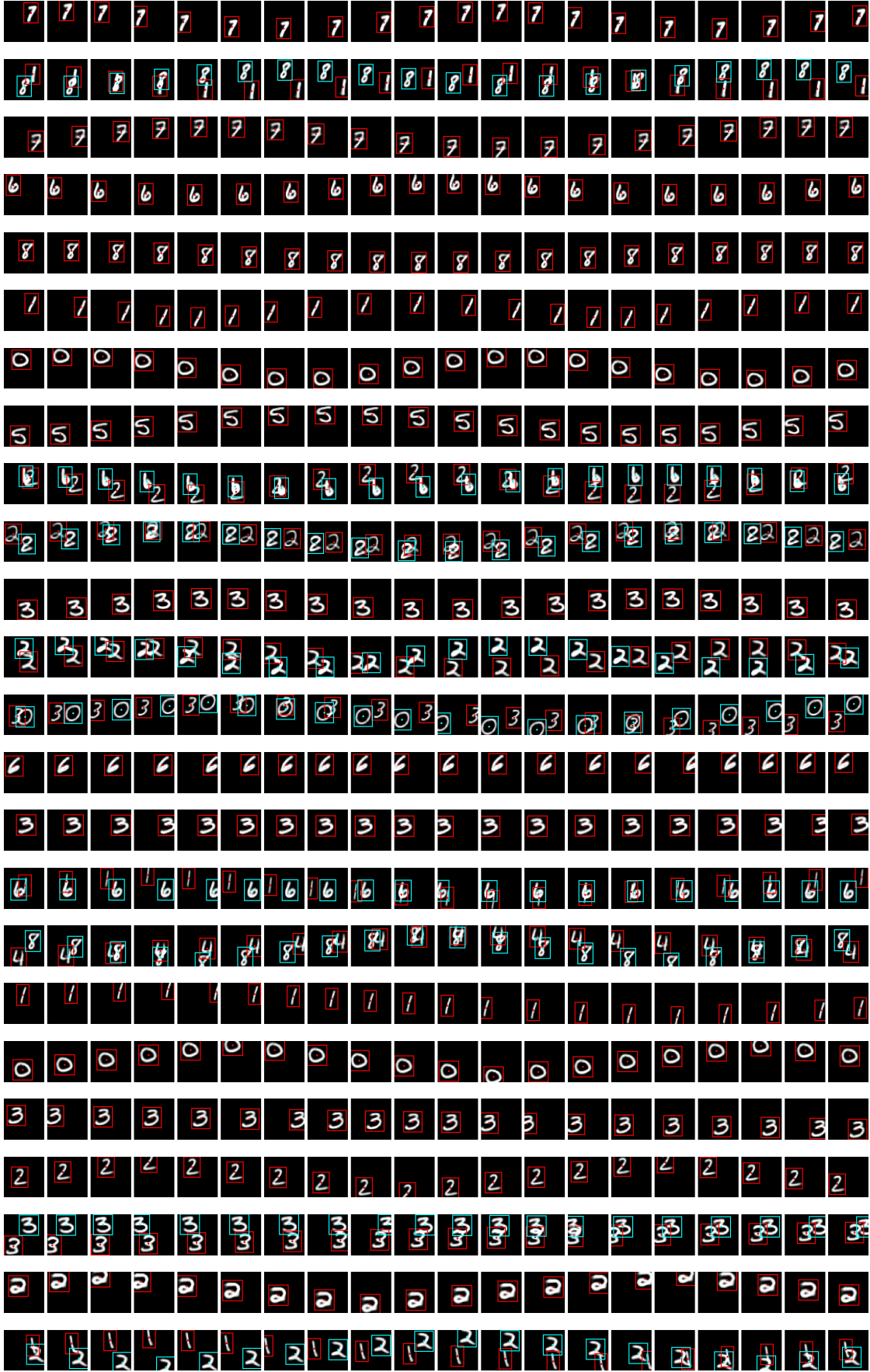}};
			}
			\end{tikzpicture}
		}
	\end{subfigure}
	\caption{
		Same as \cref{fig:vtssi-prediction}, but displaying generated frames instead of superimposing on ground truth frames.
	}
	\label{fig:vtssi-prediction-gen-frames}
\end{figure*}

 \end{document}